%% file: TTO-CONROT_Jnl_Fin.tex
\documentclass{article}

\usepackage[latin1]{inputenc}
\inputencoding{latin1}

\usepackage{float}

\usepackage{graphicx}        
\usepackage{amsmath,amsfonts} 
\usepackage{algorithmic} 
\usepackage{subfig} 
\usepackage{url} 

\usepackage[usenames,dvipsnames]{color}

\usepackage{ifthen}

\newcommand{\CHECK}{}
\newcommand{\NORMAL}{}

\usepackage{hyperref}
  \hypersetup{
        linktocpage,
        colorlinks,
        citecolor=blue,
        filecolor=blue,
        linkcolor=blue,
        urlcolor=blue,
        pdfauthor={C\'{e}sar A. Astudillo and B. John Oommen},
        pdftitle={SOMs Whose Topologies Can Be Learned With Adaptive BSTs Using Conditional Rotations}
   }

\usepackage[printonlyused]{acronym}
\usepackage{geometry}
\usepackage{setspace}
\usepackage{algorithm} 

\newcommand{\N}{\mbox{$I\!\!N$}} 

\newcommand{\R}{\mbox{$I\!\!R$}}
\newcommand{\INPUT}{\textbf{Input: }\vspace{-3mm}}
\newcommand{\OUTPUT}{\vspace{-3mm}\textbf{Output: }\vspace{-3mm}}
\newcommand{\METHOD}{\vspace{-3mm}\textbf{Method: }\vspace{-1mm}}
\newcommand{\ENDMETHOD}{\vspace{-2mm}\textbf{End Algorithm}}

\newenvironment{enumerate:alg}
{
\setlength{\leftmargini}{0.7cm}
\begin{enumerate}

  \setlength{\itemsep}{1pt}
  \setlength{\parskip}{0pt}
  \setlength{\parsep}{0pt}}
{\end{enumerate}}

\newcommand{\qed}{\nobreak \ifvmode \relax \else
      \ifdim\lastskip<1.5em \hskip-\lastskip
      \hskip1.5em plus0em minus0.5em \fi \nobreak
      \vrule height0.75em width0.5em depth0.25em\fi}

\graphicspath{{./eps/}}

\begin{document}

	\title{Self Organizing Maps Whose Topologies Can Be Learned With Adaptive Binary Search Trees Using Conditional Rotations}


\author{C\'{e}sar A. Astudillo \footnote{Universidad de Talca, Merced
437 Curic\'{o}, Chile.
\texttt{castudillo@utalca.cl}}~~\footnote{This
author is Assistant Professor at the Department of Computer Science, with the Universidad de Talca. This work is partially supported by the FONDECYT grant 11121350, Chile.
A very preliminary version of this paper was presented at AI'09,
the 2009 Australasian Joint Conference on Artificial Intelligence, Melbourne, Australia, in December 2009. That paper won the award of being the {\em Best Paper of the Conference.}
We are also very grateful for the comments made by the Associate Editor and the anonymous Referees. Their input helped in improving the quality of the final version of this paper. Thank you very much!
} \and
B. John Oommen\footnote{School of Computer Science, Carleton University, Ottawa, Canada :
K1S 5B6. \texttt{oommen@scs.carleton.ca}}~~\footnote{{\it Chancellor's Professor} ; {\it Fellow :
IEEE} and {\it Fellow : IAPR}.  This author is also an {\em Adjunct Professor} with the
University of Agder in Grimstad, Norway. The
work of this author was partially supported by NSERC, the Natural
Sciences and Engineering Research Council of Canada.}}

\date{}

\maketitle
\begin{abstract}
\CHECK
Numerous variants of \acp{SOM} have been proposed in the literature, including those which also possess an underlying structure,
and in some cases, this structure itself can be defined by the user
\NORMAL
Although the concepts of growing the \ac{SOM} and updating it have been studied, the whole issue of using a self-organizing \ac{ADS} to further enhance the properties of the underlying \ac{SOM}, has been unexplored. In an earlier work, we 
impose an \textit{arbitrary}, user-defined, tree-like topology onto the codebooks, which consequently enforced a neighborhood phenomenon and the so-called {\em tree-based} \ac{BoA}. In this paper,
we consider how the underlying tree itself can be rendered {\em dynamic} and {\em adaptively transformed}. To do this, we present methods by which a \ac{SOM} with an underlying \ac{BST} structure can be adaptively re-structured using \ac{CONROT}. These rotations on the nodes of the tree are local, can be done in constant time, and performed so as to decrease the \ac{WPL} of the entire tree. In doing this, we introduce the  pioneering concept referred to as \textit{Neural Promotion}, where neurons gain prominence in the \ac{NN} as their significance increases. We are not aware of any research which deals with the issue of Neural Promotion.
\CHECK
The advantages of such a scheme is that the user need not be aware of any of the topological peculiarities of the
stochastic data distribution.
\NORMAL
 Rather, the algorithm, referred to as the \ac{TTOCONROT}, converges in such a manner that the neurons are ultimately placed in the input space so as to represent its stochastic distribution, and additionally, the neighborhood properties of the neurons suit the best \ac{BST} that represents the data.
 These properties have been confirmed by our experimental results on a variety of data sets. We submit that all of these concepts are both novel and of a pioneering sort.

\end{abstract}

\textbf{Keywords}: \aclp{ADS}, \aclp{BST}, \aclp{SOM}

\acresetall 

\section{Introduction}
\label{sec:intro}

This paper is a pioneering attempt to merge the areas of \acp{SOM} with the theory of \acp{ADS}. Put in a  nutshell, we can describe the goal of this paper as follows: Consider a \ac{SOM} with $n$ neurons. Rather than having the neurons merely possess information about the feature space, we also attempt to \textit{link} them together by means of an underlying \ac{DS}. This \ac{DS} could be a singly-linked list, a doubly-linked list or a \ac{BST}, etc. The intention is that the neurons are governed by the laws of the \ac{SOM} \textit{and} the underlying \ac{DS}. Observe now that the concepts of ``neighborhood'' and \ac{BoA} are not based on the nearness of the neurons in the feature space, but rather on their proximity in the underlying \ac{DS}. Having accepted the above-mentioned premise, we intent to take this entire concept to a higher level of abstraction and propose to modify this \ac{DS} \textit{itself} adaptively using operations specific to it. As far as we know, the combination of these concepts has been  unreported in the literature.

Before we proceed, to place our results in the right perspective, it is probably wise to see how the concept of neighborhood has been defined in the \ac{SOM} literature.

Kohonen, in his book \cite{Kohonen1995},
mentions that
it is possible to distinguish between two basic types of neighborhood functions.
The first family involves a kernel function (which is usually of a Gaussian nature).
The second,
is the so-called neighborhood set, also known as the \acf{BoA}.
This paper focuses on the second type of neighborhood function.

Even though the traditional \ac{SOM}  is dependent on the neural distance to estimate the subset of neurons to be incorporated into the \ac{BoA}, this is not always the case for the \ac{SOM}-variants included in the literature. Indeed, the different strategies described in the state-of-the-art utilize families of schemes to define the \ac{BoA}. We mainly identify three sub-classes.

The first type of \ac{BoA} uses the concept of the neural distance as in the case of the traditional \ac{SOM}. Once the \ac{BMU} is identified, the neural distance is calculated by traversing the underlying structure that holds the neurons. An important property of the neural distance between two neurons is that it is proportional to the number of connections separating them. Examples of strategies that use the neural distance to determine the \ac{BoA} are the \ac{GCS} \cite{Fritzke1994}, the \ac{GG} \cite{Fritzke1995a}, the \ac{IGG} \cite{Blackmore1995}, the \ac{GSOM} \cite{Alahakoon2000}, the \ac{TSSOM} \cite{Koikkalainen1990}, the \ac{HFM} \cite{Miikkulainen1990}, the \ac{GHSOM} \cite{Rauber2002}, the \ac{SOTA} \cite{Dopazo1997}, the \ac{ET} \cite{Pakkanen2004}, the \ac{TTOSOM} \cite{Astudillo2011TTOSOM}, among others.

A second subset of strategies employ a scheme for determining the \ac{BoA} that does not depend on the inter-neural connections. Instead, such strategies utilize the distance in the feature space. In these cases, it is possible to distinguish between two types of \acp{NN}. The simplest situation occurs when the \ac{BoA} only considers the \ac{BMU}, i.e., it constitutes an instance of hard \ac{CL}, as in the case of the \ac{TSVQ} \cite{Koikkalainen1990} and the \ac{SOTM} \cite{Guan2006}.

A more sophisticated and computationally expensive scheme involves ranking the neurons as per their respective distances to the stimulus. In this scenario, the \ac{BoA} is determined by selecting a subset of the closest neurons. An example of a \ac{SOM} variant that uses such a ranking is the \ac{NG} \cite{Martinetz1991}.

\CHECK
According to the authors of \cite{Pakkanen2004}, the \ac{SOM}-based variants included in the literature attempt to tackle two main goals: They either try to design a more flexible topology, which is usually useful to analyze large datasets, or to reduce the
the most time-consuming task required by the \ac{SOM}, namely, the search for the \ac{BMU} when the input set has a complex nature. In this paper we focus on the former of the two mentioned goals. In other words, our goal is to enhance the capabilities of the original \ac{SOM} algorithm so as to represent the underlying data distribution and its structure in a more accurate manner. We also intend to do so by constraining the neurons so that they are related to each other, {\em not just based on their neural indices and stochastic distribution}, but also based on a \ac{BST} relationship.

Furthermore, as a long term ambition, we also anticipate methods which can accelerate
the task of locating the nearest neuron during the \ac{CL} phase.
This work will present the details of the design and implementation of how an adaptive process applied to the \ac{BST}, can be integrated into the \ac{SOM}.

Regardless of the fact that
numerous
 variants of the \ac{SOM} has been devised, few of them possess the ability of modifying the underlying topology \cite{Blackmore1995,Dittenbach2000,Dopazo1997,Fritzke1995,Guan2006,Merkl2003,Pakkanen2004,Samsonova2006}. Moreover, only a small subset use a tree as their underlying \ac{DS} \cite{Astudillo2011TTOSOM,Dittenbach2000,Dopazo1997,Guan2006,Pakkanen2004,Samsonova2006}. These strategies attempt to dynamically modify the nodes of the \ac{SOM}, for example, by adding nodes, which can be a single neuron or a layer of a \ac{SOM}-grid.
 \NORMAL
 However, our hypothesis is that it is also possible to attain to a better understanding of the unknown data distribution by performing \textit{structural} tree-based modifications on the tree, which although they preserve the general topology, attempt to modify the overall configuration, i.e., by altering the way by which nodes are \textit{interconnected}, and yet continue as a \ac{BST}. We accomplish this by dynamically adapting the edges that connect the neurons, by rotating\footnote{The operation of rotation is the one associated with \acp{BST}, as will be presently explained.} the nodes within the \ac{BST} that holds the whole structure of neurons.
  As we will explain later, this is further achieved by local modifications to the overall structure in a constant number of steps. Thus, we attempt to use rotations, tree-based neighbors \textit{and} the feature space to improve the  quality of the \ac{SOM}.

\CHECK
\subsection{Motivations}

\begin{figure}[!htp]
  \begin{center}
\subfloat[The grid learned by the \ac{SOM}.]{\label{fig:intro:a}\includegraphics[width=4cm]{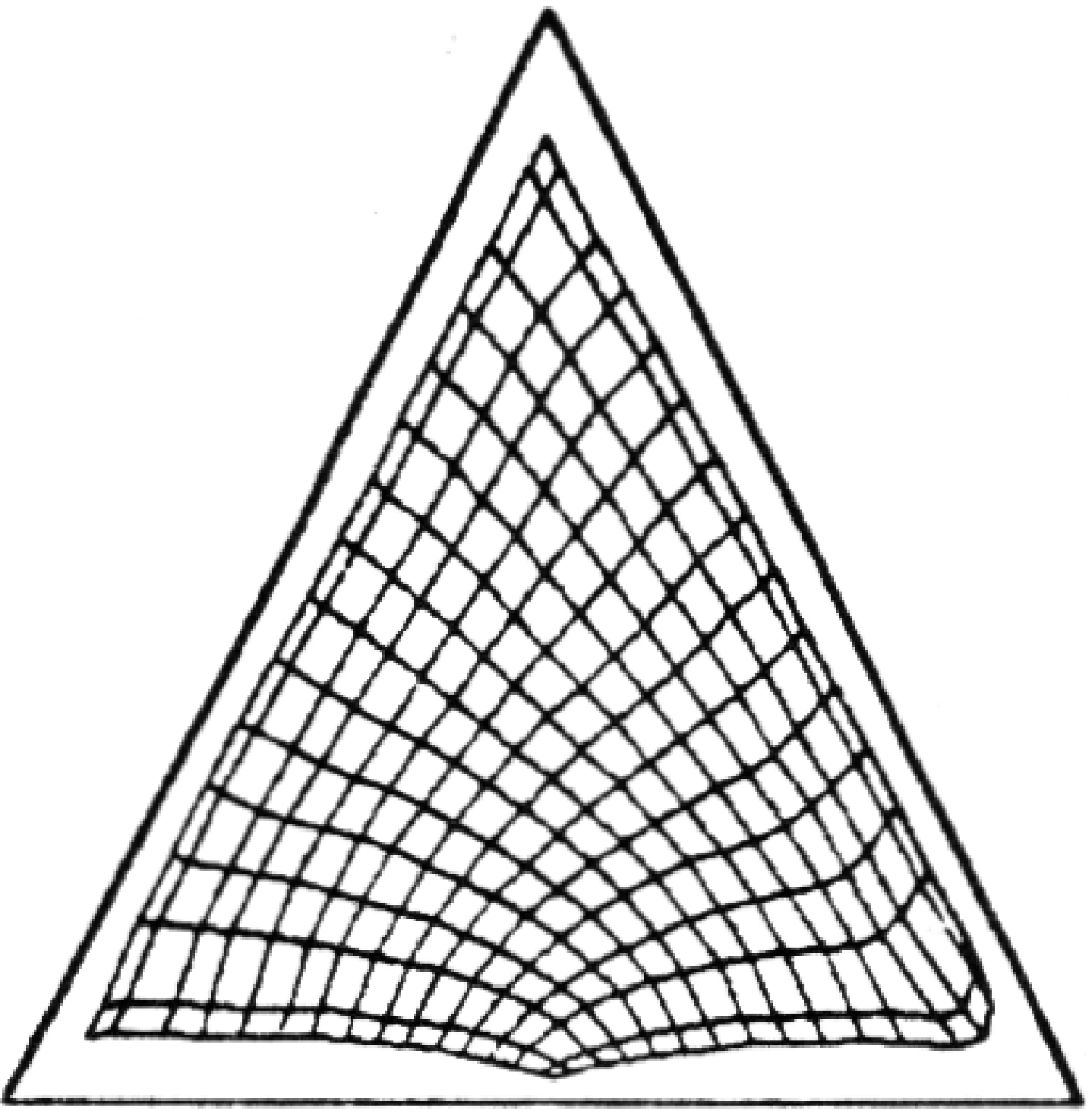}}
\hspace*{2cm}
\subfloat[The tree learned by the \ac{TTOSOM}.]{\label{fig:intro:b}\includegraphics[width=4.1cm]{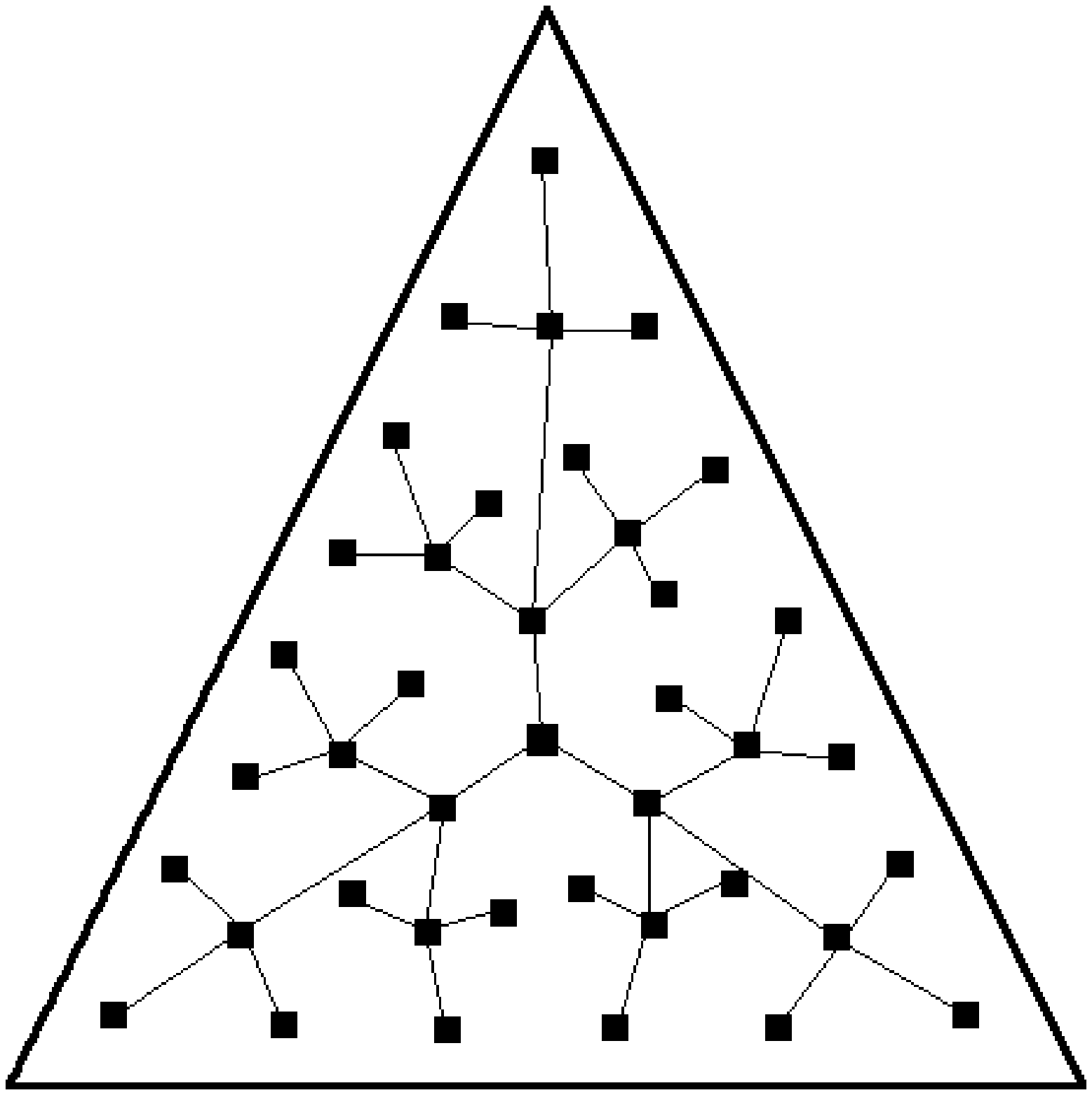}}
  \end{center}
  \caption[How a triangle-shaped distribution is learned through unsupervised learning]{How a triangle-shaped distribution is learned 
  through unsupervised learning.}
  \label{fig:intro}
\end{figure}

Acquiring information about a set of stimuli in an unsupervised manner, usually demands the deduction of its structure. In general, the \textit{topology} employed by any \ac{ANN} possessing this ability has an important impact on the manner by which it will ``absorb'' and display the properties of the input set. Consider for example, the following:
 A user may want to devise an algorithm that is capable of learning a triangle-shaped distribution as the one depicted in Figure \ref{fig:intro}. The
 \ac{SOM} tries to achieve this by defining an underlying grid-based topology and to fit the grid within the overall shape, as shown in Figure \ref{fig:intro:a} (duplicated from \cite{Kohonen1995}). However, from our perspective, a grid-like topology does not naturally fit a triangular-shaped distribution, and thus, one experiences a deformation of the original lattice during the modeling phase. As opposed to this, Figure \ref{fig:intro:b}, shows the result of applying  one of the techniques developed by us, namely the \acs{TTOSOM} \cite{Astudillo2011TTOSOM}. As the reader can observe from Figure \ref{fig:intro:b}, a 3-ary tree seems to be a far more superior choice for representing the particular shape in question.

On closer inspection, Figure \ref{fig:intro:b} depicts how the complete tree fills in the triangle formed by the set of stimuli, and further, seems to do it \textit{uniformly}. The final position of the nodes of the tree suggests that the underlying structure of the data distribution corresponds to the triangle. Additionally, the root of the tree is placed roughly in the center of mass of the triangle. It is also interesting to note that each of the three main branches of the tree, cover the areas directed towards a vertex of the triangle respectively, and their sub-branches fill in the surrounding space around them in a recursive manner, which we identify as being a holograph-like behavior.

Of course, the triangle of Figure \ref{fig:intro:b} serves only as a very simple \textit{prima facie} example to demonstrate to the reader, in an informal manner, how both techniques will try to learn the set of stimuli.
Indeed,
in real-world problems, these techniques can be employed to extract the properties of high-dimensional samples.

One can argue that imposing an initial topological configuration is not in accordance with the founding principles of unsupervised learning, the phenomenon that is supposed to occur without ``supervision'' within the human brain. As an initial response we argue that this ``supervision'' is required to enhance the \textit{training} phase, while the information we provide relates to the \textit{initialization} phase. Indeed, this is in line with the well-accepted principle \cite{Duda2000}, that very little can be automatically learned about a data distribution if no assumptions are made!

As the next step of motivating this research endeavor, we venture into a world where the neural topology \textit{and} structure are themselves learned during the training process. This is achieved by
the method that we propose in this paper, namely the \ac{TTOCONROT},
which, in essence, dynamically extends the properties of the above-mentioned \ac{TTOSOM}. Again, to accomplish this we need key concepts that are completely new to the field of \acp{SOM}, namely those related to tree-based \acf{ADS}. Indeed, as demonstrated by our experiments, the results that we have already obtained have been applauded by the research community\footnote{As mentioned earlier, a paper which reported the preliminary results of this study, won the \textit{Best Paper Award} in a well-known international \acs{AI} conference \cite{Astudillo2009a}.}, and these, to the best of our knowledge, have remained unreported in the literature.

Another reason why we are interested in such an inter-area integration, deals with the issue for devising efficient methods that add neurons to the tree. Even though the schemes that we are currently proposing in this paper focus on tree adaptation by means of rotations, we envision another type of dynamism,i.e., one which involves the expansion of the tree structure through the insertion of newly created nodes.
The state-of-the-art considers different strategies that expand trees by inserting nodes (which can be a single neuron or a \ac{SOM}-layer) that essentially are based on a \ac{QE} measure. In some of these strategies, the error measure is based on the ``hits'', i.e., the number of times a neuron has been selected as the \ac{BMU}, c.f.,
\cite{Blackmore1995,Fritzke1994,Koikkalainen1990,Pakkanen2004}.
The strategy that we have chosen for adapting the tree, namely using \ac{CONROT}, already utilizes this \ac{BMU} counter, and, distinct to the previous strategies that attempt to search for a node to be expanded (which in the case of tree-based \acp{SOM} is usually at the level of the leaves 
\cite{Koikkalainen1990,Pakkanen2004}),
we foresee and advocate a different approach.
Our \ac{TTOCONROT} method asymptotically positions frequently accessed nodes close to the root, and so, according to this property, it is the root node which should be split.
Observe that if we follow such a philosophy, one would not have to search for a node with a higher \ac{QE} measure. Rather, the \ac{CONROT}, will be hopefully, able to migrate the candidates closer to the root.
Of course, this works with assumption that a larger number of hits indicates that the degree of granularity of a particular neuron justifies refinement. The concept of using the root of the tree for growing a tree-based \ac{SOM} is, in and of itself, pioneering, as far as we know.
\NORMAL

\CHECK

\NORMAL
\subsection{Contributions of the Paper}

The contributions of the paper can be summarized as follows:

\begin{enumerate}
\item
We present an integration of the fields of \acp{SOM} and \ac{ADS}. This, we respectfully, submit as pioneering.

\item
The neurons of the \ac{SOM} are linked together using an underlying tree-based \ac{DS}, and they are governed by the laws of the \ac{TTOSOM} tree-based paradigm, and simultaneously the restructuring adaptation provided by \ac{CONROT}.
\CHECK

\item
The definition of distance between the neurons
is based on the
tree structure, and not
in the feature space. This is valid also for the \ac{BoA}, rendering the migrations distinct from the state-of-the-art.
\NORMAL

\item
The adaptive nature of the \ac{TTOCONROT} is unique because adaptation is perceived in two forms: The migration of the codebook vectors in the feature space is a consequence of the \ac{SOM} update rule, and the rearrangement of the neurons \textit{within} the tree as a result of the rotations.
\end{enumerate}


\subsection{Organization of the Paper}

The rest of the paper is organized as follows. The next section surveys the relevant literature\footnote{For the sake of space the literature review has been considerably condensed. However, given that there is no survey paper on the area of tree-based \acp{SOM} reported in the literature, we are currently preparing a paper that summarizes the field.}, which involves both the field of \acp{SOM} including their tree-based instantiations, and the respective field of \acp{BST} with conditional rotations. After that, in Section \ref{sec:tto-con-rot}, we provide an in-depth explanation of the \ac{TTOCONROT} philosophy, which is our primary contribution. The subsequent section shows the capabilities of the approach through a series of experiments, and finally, Section \ref{sec:conclusions} concludes the paper.


\section{Literature Review}


\label{sec:tto-con-rot}
\subsection{The \ac{SOM}}
One of the most important families of \acp{ANN} used to tackle
clustering problems
is the well known \ac{SOM} \cite{Kohonen1995}. Typically, the \ac{SOM} is trained using
(un)supervised learning, so as to produce a neural representation in a space whose dimension
is usually smaller than that in which the training samples lie. Further, the neurons attempt to preserve the topological properties of the input space.

\CHECK
The \ac{SOM} concentrates all the information contained in a set of $n$ input samples belonging to the $d$-dimensional space, say $\mathcal{X}=\{\textbf{x}_1,\textbf{x}_2,\ldots,\textbf{x}_n\}$, utilizing a much smaller set of  neurons, $\mathcal{C}=\{\textbf{c}_1,\textbf{c}_2,\ldots,\textbf{c}_m\}$, each of which is represented as a vector. Each of the $m$ neurons contains a weight vector $\textbf{w}=[ w_1, w_2,\ldots, w_d ]^t \in \R^d$ associated with it. These vectors are synonymously called ``weights'', ``prototypes'' or ``codebook'' vectors. The vector $\textbf{w}_i$ may be perceived as the \textit{position} of neuron $\textbf{c}_i$ in the feature space. During the training phase, the values of these weights are adjusted simultaneously so as to represent the data distribution and its structure. In each training step a stimulus (a representative input sample from the data distribution) $\textbf{x}$ is presented to the network, and the neurons compete between themselves so as to identify which is the ``winner'',
also known as the \acf{BMU}.
After identifying the \ac{BMU}, a subset of the neurons ``close'' to it are considered to be within the so-called \acf{BoA}, which further depends on a parameter specified to the algorithm, namely, the so-called radius. Thereafter, this scheme performs a migration of the codebooks within that \ac{BoA} so as to position them closer to the sample being examined. The migration factor by which this update is effected, depends on a parameter known as the learning rate, which is typically expected to be large initially, and which decreases as the algorithm proceeds, and which ultimately results in no migration at all. Algorithm \ref{alg:som} describes the details of the \ac{SOM} philosophy. In Algorithm \ref{alg:som}, the parameters are scheduled by defining a sequence $\mathcal{S}=\langle S_1,S_2,\ldots ,S_s\rangle$, where each $S_i$ corresponds to a tuple $(\eta_i,r_i,t_i)$ that specifies the learning rate, $\eta_i$, and the radius, $r_i$, for a fixed number of training steps, $t_i$. The way in which the parameters decay  is not specified in the original algorithm, and some alternatives are, e.g., that the parameters remain fixed, decrease linearly, exponentially, etc.

\begin{algorithm}
\caption{\texttt{SOM}($\mathcal{X}$,$\mathcal{S}$)} \label{alg:som}
\INPUT
\begin{description}
\item[i)]
$\mathcal{X}$, the input sample set.
\item[ii)]
$\mathcal{S}$, the schedule for the parameters.
\end{description}
\METHOD
\begin{algorithmic} [1]
\STATE  Initialize the weights $\textbf{w}_1, \textbf{w}_2,\ldots, \textbf{w}_m$, e.g., by randomly selecting elements from $\mathcal{X}$.
\REPEAT
	\STATE Obtain a sample $\textbf{x}$ from $\mathcal{X}$.
	\STATE Find the Winner neuron, i.e., the one which is most similar to $\textbf{x}$.
	\STATE Determine a subset of neurons close to the winner.
	\STATE Migrate the closest neuron and its neighbors towards $\textbf{x}$.
	\STATE Modify the learning factor and radius as per the pre-defined schedule.
\UNTIL{no noticeable changes are observed.}
\end{algorithmic}
\ENDMETHOD
\end{algorithm}

\NORMAL

Although the \ac{SOM} has demonstrated an ability to solve problems over
a wide spectrum, it possesses some fundamental drawbacks. One of these
drawbacks is that the user must specify the lattice \textit{a priori}, which has the effect that he must run the \ac{ANN} a number of times to obtain a suitable configuration. Other handicaps involve the size of the maps, where a lesser number of neurons often represent the data inaccurately.

The state-of-the-art approaches attempt to render the topology more
flexible, so as to represent complicated data distributions in a
better way and/or to make the process faster by, for instance,
speeding up the task of determining the \ac{BMU}.

There are a vast number of domain fields where the SOM has demonstrated to be useful; a compendium with all the articles that take advantage of the properties of the SOM is surveyed in \cite{Kaski1998,Oja2003}. These survey papers classify the publications related to the SOM according to their year of release. The report \cite{Kaski1998} includes the  bibliography published between the year 1981 and 1998, while the report \cite{Oja2003} includes the analogous papers published between 1998 and 2001. Further, additional recent references including the related work up to the year 2005 have been collected in a technical report \cite{Polla2009}. The more recent literature reports a host of application domains, including Medical Image Processing \cite{Akram2013}, Human Eye Detection \cite{Khosravi2008}, Handwriting Recognition \cite{Liang2012}, Image Segmentation \cite{Yao2000}, Information Retrieval \cite{Deng2007}, Object Tracking \cite{Kang2005}, etc.



\subsection{Tree-Based \acp{SOM}}
Although an important number of variants of the original \ac{SOM} have been presented through the years, we focus our attention on a specific family of enhancements in which the neurons are inter-connected using a tree topology.

The \acf{TSVQ} algorithm \cite{Koikkalainen1990} is a tree-based \ac{SOM} variant, whose topology is defined \textit{a priori} and which is static. The training first takes place at highest levels of the tree. The \ac{TSVQ} incorporates the concept of a ``frozen'' node, which implies that after a node is trained for a certain amount of time, it becomes static. The algorithm then allows subsequent units, i.e., the direct children, to be trained. The strategy utilizes a heuristic search algorithm for rapidly identifying a \ac{BMU}. It starts from the root and recursively traverses the path towards the leaves. If the unit currently being analyzed is frozen, the algorithm identifies the child which is closest to the stimulus, and performs a recursive call. The algorithm terminates when the node currently being analyzed is not a frozen node (i.e., it is currently being trained), and is returned as the \ac{BMU}.

Koikkalainen and Oja, in the same paper \cite{Koikkalainen1990} refine the idea of the \ac{TSVQ} by defining the \ac{TSSOM}, which inherits all the properties of the \ac{TSVQ}, but redefines the  search procedure and \ac{BoA}. In the case of the \ac{TSSOM}, \ac{SOM} layers of different dimensions are arranged in a pyramidal shape (which can be perceived as a \ac{SOM} with different degrees of granularity).
 It differs from the \ac{TSVQ}, in the sense that, once the \ac{BMU} is found, the direct proximity is examined to check for the \ac{BMU}. On the other hand, the \ac{BoA} differs in that, instead of considering only the \ac{BMU}, its direct neighbors (in the pyramid) will also be considered.

The \acf{SOTA} \cite{Dopazo1997} is a dynamically growing tree-based \ac{SOM} which, according to their authors, take some analogies from the \acf{GCS} \cite{Fritzke1994}. The \ac{SOTA} utilizes a binary tree as the underlying structure, and similarly to other strategies (e.g., the \ac{TSSOM} \cite{Koikkalainen1990} and the \acf{ET} \cite{Pakkanen2004} explained below), it considers the migration of the neurons \textit{only} if they correspond to leaf nodes within the tree structure. Its \ac{BoA} depends on the neural tree and is defined for two cases. The most general case occurs when the parent of the \ac{BMU} is not the root, i.e., a situation in which the \ac{BoA} is composed by the \ac{BMU}, its sibling and its parent node. Otherwise, the \ac{BoA} constitutes the \ac{BMU} \textit{only}. The \ac{SOTA} triggers a  growing mechanism that utilizes a \ac{QE} to determine the node to be split into two new descendants.

In \cite{Dittenbach2000} the authors presented a tree-based \ac{SOM} called the \acf{GHSOM}, in which each node corresponds to an independent \ac{SOM}. 
The expansion of the structure is dual: The first type of adaptation is conceived by inserting new rows (or columns) to the \ac{SOM} grid that is currently being trained, while the second type is implemented by adding layers to the hierarchical structure.
Both types of dynamism depend on the verification of \ac{QE} measures.


\CHECK
The \ac{SOTM} \cite{Guan2006} is a tree-based \ac{SOM} which is also inspired by the \ac{ART} \cite{Carpenter1988}. In the \ac{SOTM}, when the input is within a threshold distance from the \ac{BMU}, the latter is migrated.  Otherwise, a new neuron is added to the tree. Thus, in the \ac{SOTM}, the subset of neurons to be migrated depends only on the distance in the \textit{feature} space, and not in the neural distance, as most of the tree-based \ac{SOM} families.
\NORMAL

In \cite{Pakkanen2004}, the authors have proposed a tree-structured \ac{NN} called the \acf{ET}, which takes advantage of a sub-optimal procedure
\CHECK
(adapted from the one utilized by the \ac{TSVQ})
\NORMAL
to identify the \ac{BMU} in $O(\log |V|)$ time, where $V$ is the set of neurons. The \ac{ET} adds neurons dynamically, and incorporates the concept of a ``frozen'' neuron (explained above), which is a non-leaf node that does not participate in the training process, and which is thus removed from the \ac{BoA}. Similar to the \ac{TSVQ}, the training phase terminates when all the nodes become frozen.

\CHECK
The \acf{TTOSOM} \cite{Astudillo2011TTOSOM}, which is central to this paper,
is a tree-based \ac{SOM} in which each node can possess an arbitrary number of children. Furthermore, it is assumed that the user has the ability to describe/create such a tree whose topological configuration is preserved through the training process.
The \ac{TTOSOM} uses a particular \ac{BoA} that includes nodes (leaf and non-leaf ones) that are within a certain neural distance (the so-called ``radius'').
\NORMAL
An interesting property displayed by this strategy is its ability to reproduce the results obtained by Kohonen, when the nodes of the \ac{SOM} are arranged linearly, i.e., in a list. In this case, the \ac{TTOSOM} is able to adapt this 1-dimensional grid to a 2-dimensional (or multi-dimensional) object in the same way as the \ac{SOM} algorithm does \cite{Astudillo2011TTOSOM}. This was a phenomenon that was not possessed by  prior hierarchical \ac{SOM}-based networks reported in the literature\footnote{
\CHECK
The \ac{SOM} possesses the ability to learn the data distribution by utilizing a unidimensional topology \cite{Kohonen1995}, i.e., the neighbors are defined along a grid in each direction. Further, when this is the case, one can encounter that the  unidimensional topology forms a so-called \textit{Peano} curve \cite{Peano1890}. The \ac{TTOSOM} also possesses this interesting property, when the tree topology is linear. The details of how this is achieved is presented in detail in  \cite{Astudillo2011TTOSOM}, including the explanation of how other tree-based techniques fail to achieve this task.
\NORMAL
}.
Additionally, if the original topology of the tree followed the overall shape of the data distribution, the results reported in \cite{Astudillo2011TTOSOM} (and also depicted in the motivational section) showed that is also possible to obtain a symmetric topology for the codebook vectors.
In a more recent work \cite{Astudillo2013}, the authors have enhanced the TTOSOM to perform classification in a semi-supervised fashion.  The method presented in \cite{Astudillo2013} first learns the data distribution in an unsupervised manner. Once labeled instances become available, the clusters are labeled using the evidence. According to the results presented in \cite{Astudillo2013}, the number of neurons required to accurately predict the category of novel data are only a small portion of the cardinality of the input set.





\section{Merging ADS and TTOSOM}
\label{sec:adaptive}
\subsection{\acfp{ADS} for \acp{BST}}
\label{sec:ads}
One of the primary goals of the area of \ac{ADS} is to achieve an optimal arrangement of the elements, placed at the nodes of the structure, as the number of iterations increases. This reorganization can be perceived to be both automatic and adaptive, such that on convergence, the \ac{DS} tends towards an optimal configuration with a minimum average access time. In most cases, the most probable element will be positioned at the root (head) of the tree (\ac{DS}), while the rest of the tree is recursively positioned in the same manner. The solution to obtain the \textit{optimal} \ac{BST} is well known when the access probabilities of the nodes are known \textit{a priori} \cite{Knuth1998}. However, our research concentrates on the case when these access probabilities are \textit{not known a priori}.
\CHECK
In this setting, one effective solution is due to Cheetham \textit{et al.} and uses the concept of \ac{CONROT} \cite{Cheetham1993},
which reorganizes
the \ac{BST} so as to asymptotically produce the optimal form.
Additionally, unlike most of the algorithms that are otherwise reported in the literature, this move is not done on every data access operation -- it is performed if and only if the overall \ac{WPL} of the resulting \ac{BST} decreases.


A \ac{BST} may be used to store records
whose keys are members of an ordered set
$\mathcal{A} = \{A_1, A_2, \ldots, A_N\}$. The records are stored in such a
way that a symmetric-order traversal of the tree will yield the
records in an ascending order.
If we are given $\mathcal{A}$ and the set of access probabilities $\mathcal{Q} =
\{Q_1, Q_2, \ldots, Q_N\}$, the problem
of constructing efficient \acp{BST} has
been extensively studied. The optimal algorithm due to Knuth \cite{Knuth1998}, uses
dynamic programming and produces the optimal
\ac{BST} using $O(N^2)$ time and space.
In this paper, we consider the scenario in which $\mathcal{Q}$, the access probability vector, is not known \textit{a
priori}. We seek a scheme which dynamically rearranges itself
and asymptotically generates a tree which minimizes the access cost of the keys.



\begin{figure}[!ht]
  \centering
  \subfloat[The tree before a rotation is performed. The contents of the nodes are their data values, which in this case are the characters $\{a,b,c,d,e\}$.\label{fig:rot:a}]{
    \includegraphics[scale=1.5]{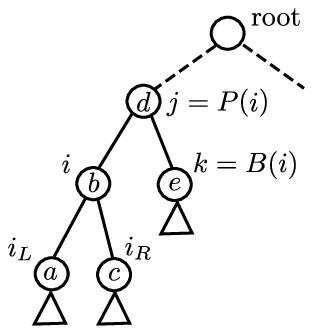}
  }
  \hspace{15mm}
  \subfloat[The tree after a rotation is performed on node $i$.\label{fig:rot:b}]{
    \includegraphics[scale=1.5]{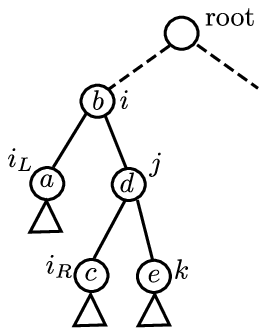}
  }
  \caption{The \ac{BST} before and after a Rotation is performed.}\label{fig:rot}
\end{figure}

The primitive tree restructuring operation used in most \ac{BST}
schemes is the well known operation of Rotation \cite{Adelson1962}.
We describe this operation as follows.
Suppose that there exists a node $i$ in a \ac{BST}, and that it has a
parent node $j$, a left child, $i_L$, and a right child, $i_R$.
\CHECK
The function $P(i)=j$ relates node $i$ with its parent $j$ (if it exists). Also, let $B(i)=k$ relate node $i$ with its sibling $k$, i.e., the node (if it exists) that shares the same parent as $i$.
\NORMAL
Consider
the case when $i$ is itself a left child (see Figure \ref{fig:rot:a}). A rotation is
performed on node $i$ as follows: $j$ now becomes the right child,
$i_R$ becomes the left child of node $j$, and all the other nodes remain
in their same relative positions (see Figure \ref{fig:rot:b}). The case when
node $i$ is a right child is treated in a symmetric manner. This operation
has the effect of raising (or promoting) a specified node in the tree structure
while preserving the lexicographic order of the elements (refer
again to Figure\ref{fig:rot:b}).


A few memory-less tree reorganizing schemes\footnote{This review is necessary brief. A more detailed version is found in \cite{Cormen2001,Lai1990}.} which use this operation have been presented in the literature among which are the Move-to-Root and the simple Exchange rules \cite{Allen1978}.
%
%
In
the Move-to-Root Heuristic,
 each time a record is accessed, rotations are performed on it
 in an upwards direction until it becomes the root of the tree.
 On the other hand,
 the simple Exchange rule
 rotates the accessed element
 one level towards the root.
%

Sleator and Tarjan \cite{Sleator1985}
introduced a technique, which also
moves the accessed record up to the root of the tree using a
restructuring operation called ``Splaying'', which actually is a multi-level
generalization of the rotation.
 Their structure, called the Splay
Tree, was shown to have an amortized time complexity of
$O(\log N)$ for a complete set of tree operations which included
insertion, deletion, access, split, and join.

 The literature also records various schemes which adaptively
 restructure the tree with the aid of additional memory
 locations.
 Prominent among them is the \ac{MT} \cite{Bitner1979} and Mehlhorn's \ac{DT} \cite{Mehlhorn1979}.
 The \ac{MT} is a dynamic version of a tree structuring
 method originally suggested by Knuth \cite{Knuth1998}.

In spite of all their advantages, all of the schemes mentioned
above have drawbacks, some of which are more serious than
others.
The memory-less schemes have one major disadvantage,
which is that both the Move-to-Root and Splaying
rules always move the accessed record up to the root of
the tree. This means that if a nearly-optimal arrangement
is reached, a single access of a seldomly-used record will
disarrange the tree along the entire access path, as the element
is moved upwards to the root.

As opposed to these schemes, the \ac{MT} rule
does not move the accessed element to the root every time.
But, as reported in \cite{Cheetham1993}, in practice, it does not
perform well.
The weakness of the \ac{MT}
lies in the fact that it considers only the frequency
counts for the records,
which leads to
the undesirable property that a single rotation may move a subtree with a
relatively large probability weight downwards,
thus increasing the cost of the tree.


This paper uses a particular heuristic, namely, the \ac{CONROT-BST} \cite{Cheetham1993},
 which has been shown to reorganize
 a \ac{BST} so as to asymptotically arrive at an optimal form.
 In its optimized version, the scheme, referred to Algorithm \ac{CONROT-BST}, requires the maintenance of a single memory location per record, which
keeps track of
the number of accesses to the subtree rooted at
 that record.
The \ac{CONROT-BST} algorithm specifies
how an accessed
 element can be rotated towards the root of the tree so as to
 minimize the overall cost of the entire tree. Finally, unlike most of
 the algorithms that are currently in the literature, this move is
not done on every data access operation. It is performed if and
only if the overall \ac{WPL} of the resulting \ac{BST} decreases.
In essence Algorithm \ac{CONROT-BST} attempts to minimize the \ac{WPL} by incorporating the
statistical information about the accesses to the various nodes
\textit{and subtrees} rooted at the corresponding nodes.

The basic condition for the rotation of a node is that the \ac{WPL} of
the entire tree must decrease as a result of a single rotation. This is achieved by a  so-called \textit{Conditional} Rotation.
To define the concept of a Conditional Rotation, we define $\tau_i(n)$ as the total number of
accesses to the subtree rooted at node $i$.
One of the biggest advantages of the \ac{CONROT-BST} heuristic is that it only requires the maintenance and processing of the values stored at a specific node and its direct neighbors, i.e., its parent and both children, if they exist.

Algorithm \ac{CONROT-BST}, formally given in Algorithm \ref{alg:cond-rot-bst}, describes the process of the conditional rotations for a \ac{BST}. The algorithm receives two parameters, the first of which corresponds to a pointer to the root of the tree, and the second which corresponds to the key to be searched, which is assumed to be present in the tree. When a node access is requested, the algorithm seeks for the node from the root down towards the leaves.

{\small
    \begin{algorithm}
    \caption{\texttt{CONROT-BST}($j$,$k_i$)}
    \label{alg:cond-rot-bst}
    \INPUT
    \begin{description}
	    \item[i)]
		    $j$, A pointer to the root of a binary search tree $T$
		    \vspace{-3mm}
	    \item[ii)]
		    $k_i$, A search key, assumed to be in $T$
    \end{description}
    \OUTPUT
    \begin{description}
	    \item[i)]
		    The restructured tree $T'$
		    \vspace{-3mm}
	    \item[ii)]
		    A pointer to the record $i$ containing $k_i$
    \end{description}
    \METHOD
    \begin{algorithmic} [1]
    \STATE $\tau_j \leftarrow \tau_j + 1$
    \IF{ $k_i = k_j$ }
	    \IF{ \texttt{is-left-child}($j$) = \texttt{TRUE} }  
		    \STATE $\Psi_j \leftarrow 2\tau_j - \tau_{jR} - \tau_{P(j)}$
	    \ELSE
		    \STATE $\Psi_j \leftarrow 2\tau_j - \tau_{jL} - \tau_{P(j)}$
	    \ENDIF
	    \IF{  $\Psi_j > 0$ }  
		    \STATE \texttt{rotate-upwards}($j$)
		    \STATE \texttt{recalculate-tau}($j$)
		    \STATE \texttt{recalculate-tau}($P(j)$)
	    \ENDIF
	    \STATE \textbf{return} record $j$
    \ELSE
	    \IF{  $k_i < k_j$ }  
		    \STATE \texttt{CONROT-BST}(  \texttt{left-child}($j$) , $k_i$ )
	    \ELSE
		    \STATE \texttt{CONROT-BST}( \texttt{right-child}($j$) , $k_i$ )
	    \ENDIF
    \ENDIF
    \end{algorithmic}
    \ENDMETHOD
    \end{algorithm}
}

The first task accomplished by the Algorithm \ac{CONROT-BST} is the updating of the counter  $\tau$ for the present node along the path traversed. After that, the next step consists of determining whether or not the node with the requested key has been found.
\CHECK
When this occurs, the quantities defined by Equations \eqref{eq:psi-l} and \eqref{eq:psi-r} are computed to determine the value of a quantity referred to as $\Psi$, where:

    \begin{equation}
    \Psi_j = 2\tau_j - \tau_{jR} - \tau_{P(j)}
    \label{eq:psi-l}
    \end{equation}

\noindent
when $j$ is the left child of its parent, $P(j)$, and

    \begin{equation}
    \Psi_j = 2\tau_j - \tau_{jL} - \tau_{P(j)}
    \label{eq:psi-r}
    \end{equation}

\noindent
when $j$ is a right descendant of $P(j)$.
\NORMAL

When $\Psi$ is less than zero, an upward rotation is performed. The authors of \cite{Cheetham1993} have shown that this single rotation leads to a decrease in the overall \ac{WPL} of the \textit{entire} tree. This occur in line 9 of the algorithm, in which the method \texttt{rotate-upwards} is invoked. The parameter to this method is a pointer to the node $j$. The method does the necessary operations required to rotate the node upwards, which means that if the node $j$ is the left child of the parent, then this is equivalent to performing a right rotation over $P(j)$, the parent of $j$. Analogously, when $j$ is the right child of its parent, the parent of $j$ is left-rotated instead. Once the rotation takes place, it is necessary to update the corresponding counters, $\tau$. Fortunately this task only involve the updating of $\tau_i$, for the rotated node, and the counter of its parent, $\tau_{P(i)}$. The last part of the algorithm, namely lines 14--19, deals with the further search for the key, which in this case is achieved recursively.

The reader will observe that all the tasks invoked in the algorithm are performed in constant time, and in the worst case, the recursive call is done from the root down to the leaves, leading to a $O(h)$ running complexity, where $h$ is the height of the tree.

\subsection{The \acf{TTOCONROT}}
This section concentrates on the details of the integration between the fields of \ac{ADS} and the \ac{SOM}, and in particular, the \ac{TTOSOM}. Although merging \ac{ADS} and the \ac{SOM} is relevant to a wide spectrum of \acp{DS}, we focus our scope by considering only tree-based structures. More specifically we shall concentrate on the integration of the \ac{CONROT-BST} heuristic \cite{Cheetham1993} into a \ac{TTOSOM} \cite{Astudillo2011TTOSOM}, both of which were explained in the preceding sections.

\CHECK
We can conceptually distinguish our method, namely, the \acl{TTOSOM} with \acl{CONROT} (\acused{TTOCONROT}\ac{TTOCONROT}) from
its components and properties.

In terms of components, we detect five elements. First of all, the \ac{TTOCONROT} has a set of neurons, which, like all \ac{SOM}-based methods, represents the data space in a condensed manner.  Secondly, the \ac{TTOCONROT} possesses a connection between the neurons, where the neighbor of any specific neuron is based on a  nearness measure that is tree-based.   The third and fourth components involve the migration of the neurons. Similar to the reported families of \acp{SOM}, a subset of neurons closest to the winning neuron are moved towards the sample point using a \ac{VQ} rule. However, unlike the reported families of \acp{SOM}, the identity of the neurons that are moved is based on the tree-based proximity and not on the feature-space proximity. Finally, the \ac{TTOCONROT} possesses tree-based mutating operations, namely the above-mentioned conditional rotations.

With respect to the properties of the \ac{TTOCONROT}, we mention the following. First of all, it is adaptive, with regard to the migration of the points.  Secondly, it is also adaptive with regard to the identity of the neurons moved. Thirdly, the distribution of the neurons in the feature space mimics the distribution of the sample points. Finally, by virtue of the conditional rotations, the entire tree is optimized with regard to the overall accesses, which is a unique phenomenon (when compared to the reported family of \acp{SOM}) as far as we know.

As mentioned in the introductory section, the general dynamic adaptation of \ac{SOM} lattices reported in the literature considers essentially adding (and in some cases deleting) nodes/edges.
However the concept of modifying the underlying structure's \textit{shape} itself has been unrecorded. Our hypothesis is that this is advantageous by means of a repositioning of the \textit{nodes} and the consequent \textit{edges}, as seen when one performs rotations on a \ac{BST}. In other words, we place our emphasis on the self-arrangement which occurs as a result of restructuring the \ac{DS} representing the \ac{SOM}. In this case, as alluded to earlier, the restructuring process is done between the connections of the neurons so as to attain an asymptotically optimal configuration, where nodes that are accessed more frequently will tend to be placed close to the root. We thus obtain a new species of tree-based \acp{SOM} which is self-arranged by performing rotations \textbf{conditionally}, \textbf{locally} and in a \textbf{constant number of steps}.
\NORMAL

%
The primary goal of the field of \ac{ADS} is to have the structure and its elements attain an optimal configuration as the number of iterations increases. Particularly, among the \acp{ADS} that use trees as the underlying topology, the common goal is to minimize the overall access cost, and this roughly means that one places the most frequently accessed nodes close to the root, which is also what \ac{CONROT-BST} moves towards. Although such an adaptation can be made on any \ac{SOM} paradigm, the \ac{CONROT} is relevant to a tree structure, and thus to the \ac{TTOSOM}. This further implies that some specific settings/modifications must be applied to achieve the integration between the two paradigms.

We start by defining a \ac{BSTSOM} as a special instantiation of a \ac{SOM} which uses a \ac{BST} as the underlying topology. An \ac{ABSTSOM} is a further refinement of the \ac{BSTSOM} which, during the training process,  employs a technique that automatically modifies the configuration of the tree. The goal of this adaptation is to facilitate and enhance the search process. This assertion must be viewed from the perspective that for a \ac{SOM}, neurons that represent areas with a higher density, will be queried more often.

Every \ac{ABSTSOM} is characterized by the following properties. First, it is \textbf{adaptive}, where, by virtue of the \ac{BST} representation this adaptation is done by means of rotations, rather than by merely deleting or adding nodes. Second, the neural network corresponds to a \ac{BST}. The goal is that the \ac{NN} maintains the essential stochastic and topological properties of the \ac{SOM}.

\subsubsection{Neural Distance}


As in the case of the \ac{TTOSOM} \cite{Astudillo2011TTOSOM}, the \textit{Neural Distance}, $d_N$, between two neurons depends on the \textit{number} of unweighted connections that separate them in the user-defined tree. It is consequently the number of edges in the shortest path that connects the two given nodes. More explicitly, the distance between two nodes in the tree, is defined as the minimum number of edges required to go from one to the other.
In the case of trees, the fact that there is only a \textit{single} path connecting two nodes implies the uniqueness of the shortest path, and permits the efficient calculation of the distance between them by a node traversal algorithm. Note however, that in the case of the \ac{TTOSOM}, since the tree itself was \emph{static}, the inter-node distances can be pre-computed \emph{a priori}, simplifying the computational process. The situation changes when the tree is dynamically modified as we shall explain below.

The implications of having the tree which describes the \ac{SOM} to be dynamic, are three-fold. First of all, the siblings of any given node may change at every time instant. Secondly, the parents and ancestors of the node under consideration could also change at every instant. But most importantly, the structure of the tree itself could change, implying that nodes that were neighbors at any time instant may not continue to be neighbors at the next. Indeed, in the extreme case, if a node was migrated to become the root, the fact that it had a parent at a previous time instant is irrelevant at the next. This, of course, changes the entire landscape, rendering the resultant \ac{SOM} to be unique and distinct from the state-of-the-art. An example will clarify this.

\begin{figure}[!ht]
  \centering
  \subfloat[\label{fig:before-rotation}]{
    \includegraphics[width=5.5cm]{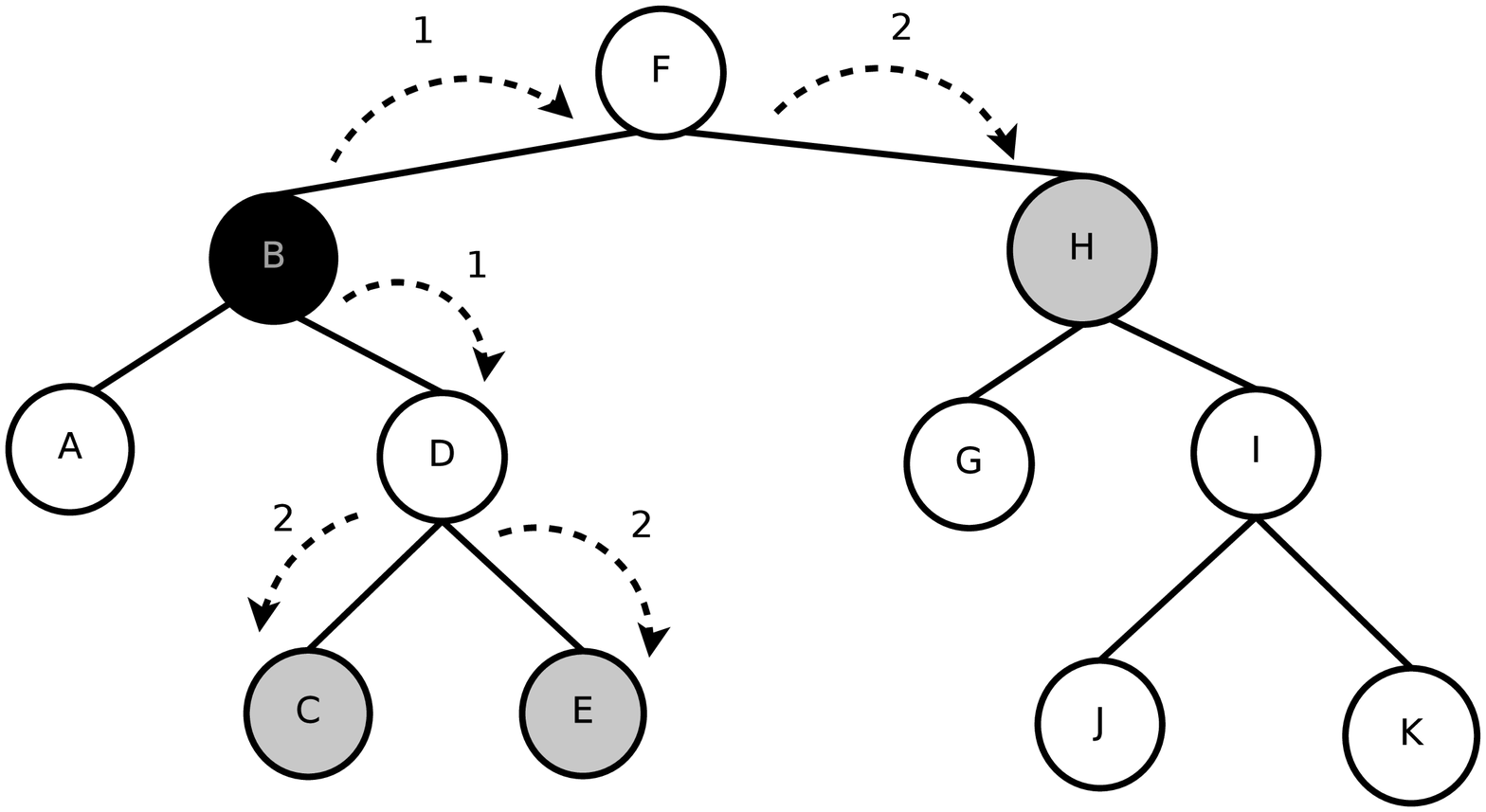}
  }
  \hspace{2mm}
  \subfloat[\label{fig:the-rotation}]{
    \includegraphics[width=5.5cm]{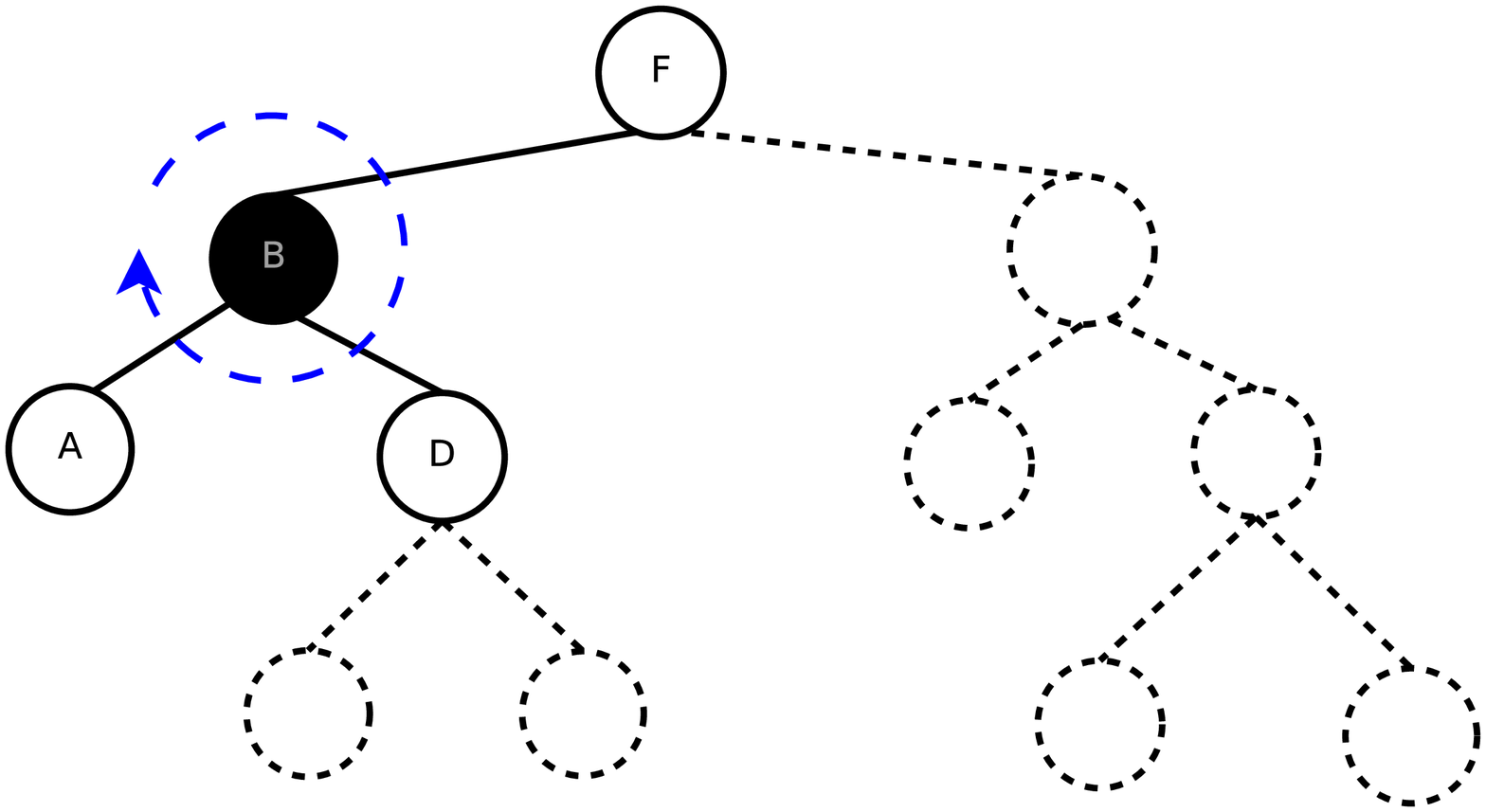}
  }\\
\hspace{2mm}
  \subfloat[\label{fig:the-rotation2}]{
    \includegraphics[width=5.5cm]{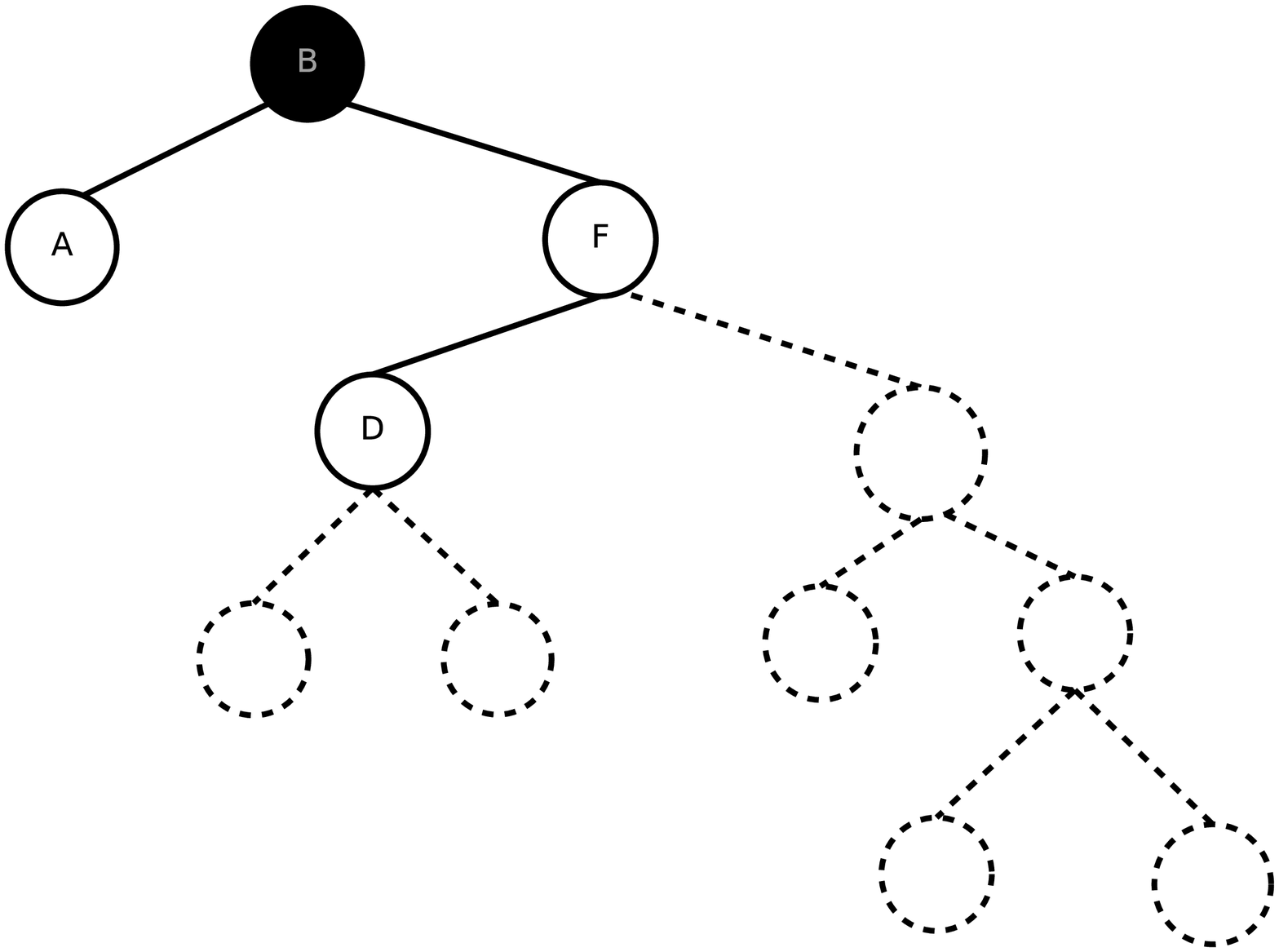}
  }
\hspace{2mm}
  \subfloat[\label{fig:after-rotation}]{
    \includegraphics[width=5.5cm]{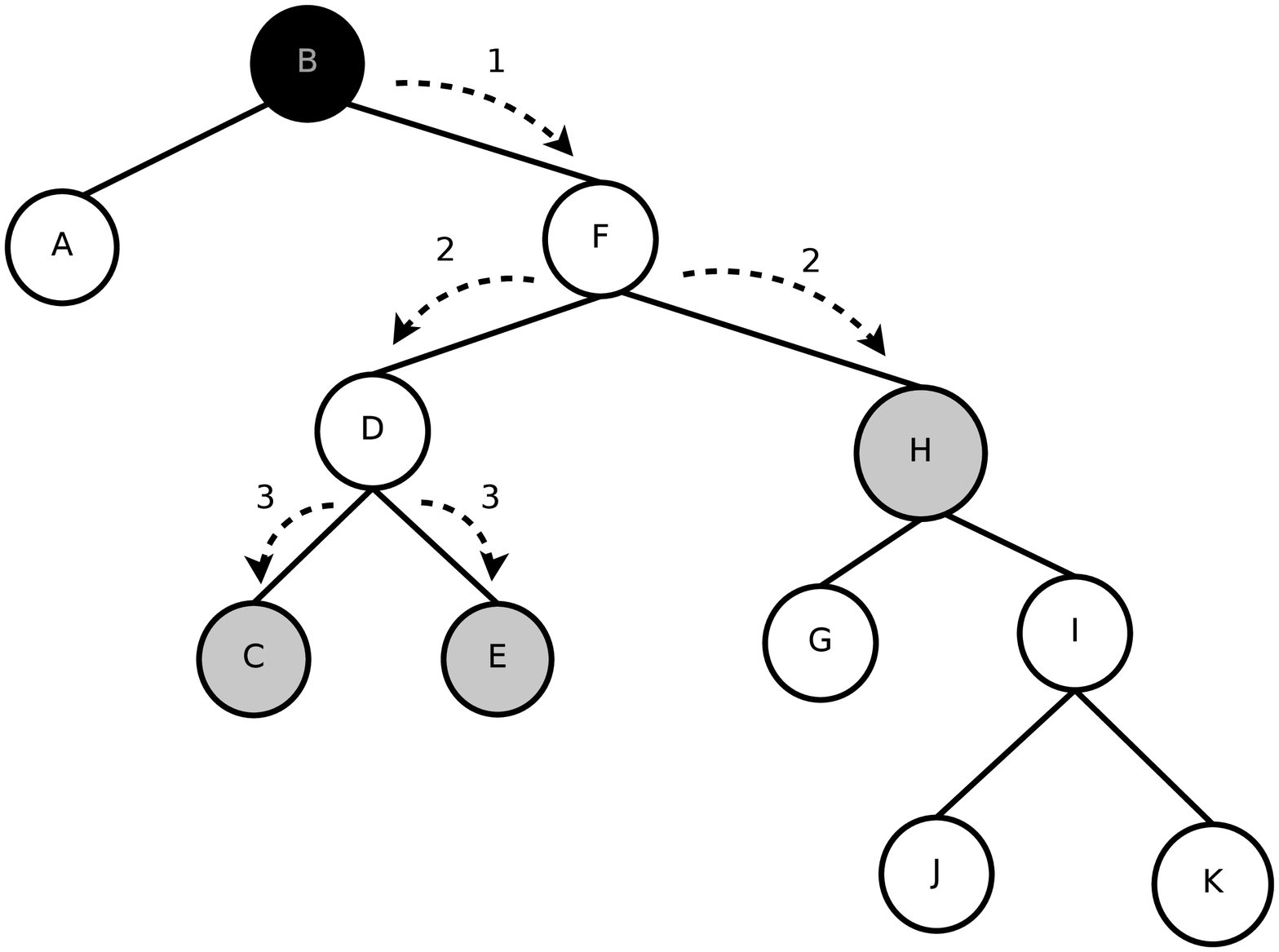}
  }

    \caption[Example of the neural distance before and after a rotation.]{ Example of the neural distance before and after a rotation. In Figure \ref{fig:before-rotation} nodes $H$, $C$ and $E$ are equidistant from $B$ even though they are at different levels in the tree. Figures \ref{fig:the-rotation} and \ref{fig:the-rotation2} show the process of rotating node A upwards. Finally, Figure \ref{fig:after-rotation} depicts the state of the tree after the rotation when only $C$ and $E$ are equidistant from $B$, and their distance to $B$ has increased by unity. On the other hand, although $H$ has changed its position, its distance to $B$ remains the same.
}
  \label{fig:TTOROT-neural-distance}
\end{figure}

Consider Figure \ref{fig:TTOROT-neural-distance}, which illustrates the computation of the neural distance for various scenarios. First, in Figure \ref{fig:before-rotation}, we present the scenario when the node accessed is $B$. Observe that the distances are depicted with dotted arrows, with an adjacent numeric index specifying the current distance from node $B$. In the example, prior to an access, nodes $H$, $C$ and $E$ are all at a distance of $2$ from node $B$, even though they are at different levels in the tree.  The reader should be aware that non-leaf nodes may also be involved in the calculation, as in the case of node $H$. Figures \ref{fig:the-rotation} and \ref{fig:the-rotation2} show the process when node $B$ is queried, which in turn triggers a rotation of node $B$ upwards. Observe that the rotation itself only requires local modifications, leaving the rest of the tree untouched. For the sake of simplicity and explicitness, unmodified areas of the tree are represented by dashed lines. Finally, Figure \ref{fig:after-rotation} depicts the configuration of the tree after the rotation is performed. At this time instant, $C$ and $E$ are both at distance of 3 from $B$, which means that they have increased their distance to $B$ by unity. Moreover, although node $H$ has changed its position, its distance to $B$ remains unmodified. Clearly, the original distances are not necessarily preserved as a consequence of the rotation.

Generally speaking, there are four regions of the tree that remain unchanged. These are, namely, the portion of the tree above the parent of the node being rotated, the portion of tree rooted at the right child of the node being rotated, the portion of tree rooted at the left child of the node being rotated, and the portion of tree rooted at the sibling of the node being rotated. Even though these four regions remain unmodified, the neural distance in these regions are affected, because the rotation could lead to a modification of the distances to the nodes.

Another consequence of this operation that is worth mentioning is the following: The distance between any two given nodes that belong to the same unmodified region of the tree is preserved after a rotation is performed. The proof of this assertion is obvious, inasmuch as the fact remains that every path between nodes in any unmodified sub-tree remains with the same sub-tree. This property is interesting because it has the potential to accelerate the computation of the respective neural distances.

\subsubsection{The \acl{BoA}}

A concept closely related to the neural distance, is the one referred to as the \acf{BoA} which is the subset of nodes within a distance of $r$ away from the node currently examined. Those nodes are in essence those which are to be migrated toward the signal presented to the network. This concept is valid for all \ac{SOM}-like \acp{NN}, and in particular for the \ac{TTOSOM}. We shall now consider how this bubble is modified in the context of rotations. The concept of the bubble involves the consideration of a quantity, the so-called \textit{radius}, which establishes how big the \ac{BoA} is, and which therefore has a direct impact on the number of nodes to be considered. The \ac{BoA} can be formally defined as \cite{Astudillo2011TTOSOM}

\begin{equation}
 B(v_i;T,r)=\lbrace v | d_N(v_i,v;T) \leq r \rbrace,
 \label{eq:bubble-static}
\end{equation}

\noindent where $v_i$ is the node currently being examined, and $v$ is an arbitrary node in the tree $T$, whose nodes are $V$. Note that $B(v_i,T,0)=\{v_i\}$, $B(v_i,T,i) \supseteq B(v_i,T,i-1)$ and $B(v_i,T,|V|)=V$ which generalizes the special case when the tree is a (simple) directed path.

\begin{algorithm}
\caption{\texttt{TTOSOM\_Calculate\_Neighborhood}($B$,$v$,$r$)} \label{calculateneighborhood}
\INPUT
\begin{description}
\item[i)]
 $B$, the set of the nodes in the bubble of activity identified so far.
 \vspace{-3mm}
\item[ii)]
$v$, the node from where the bubble of activity is calculated.
\vspace{-3mm}
\item[iii)]
$r$, the current radius of the bubble of activity.
\end{description}
\OUTPUT
\begin{description}
\item[i)]
 The set of nodes in the bubble of activity.
\end{description}
\METHOD
\begin{algorithmic}[1]
\label{alg::cn}
\IF{$r\leq 0$}
    \STATE \textbf{return}
\ELSE
	\FORALL{\textit{child} $\in$ $v$.\texttt{getChildren}()}
		\IF{\textit{child} $\notin B$}
    			\STATE $B \leftarrow B + \{$\textit{child}$\}$
        		\STATE \texttt{TTOSOM\_Calculate\_Neighborhood}$(B,$\textit{child}$,r-1)$
    		\ENDIF
	\ENDFOR
	\STATE \textit{parent}=$v$.\texttt{getParent}();
	\IF{\textit{parent} $\neq$ \texttt{NULL} and  \textit{parent} $\notin$ $B$}
		\STATE $B \leftarrow B + \{$\textit{parent}$\}$
		\STATE \texttt{TTOSOM\_Calculate\_Neighborhood}$(B,$\textit{parent}$,r-1)$
   	\ENDIF

\ENDIF
\end{algorithmic}
\ENDMETHOD
\end{algorithm}

To clarify how the bubble changes in the context of rotations, we first describe the context when the tree is static. As presented in \cite{Astudillo2011TTOSOM}, the function \texttt{TTOSOM\_Calculate\_Neighborhood} (see Algorithm \ref{calculateneighborhood}) specifies the steps involved in the calculation of the subset of neurons that are part of the neighborhood of the \ac{BMU}. This computation involves a collection of parameters, including $B$, the current subset of neurons in the proximity of the neuron being examined, $v$, the \ac{BMU} itself, and $r \in \N$ the current radius of the neighborhood. When the function is invoked for the first time, the set $B$ contains only the \ac{BMU} marked as the current node, and through a recursive call, $B$ will end up storing the entire set of units within a radius $r$ of the \ac{BMU}. The tree is recursively traversed for all the direct topological neighbors of the current node, i.e., in the direction of the direct parent and children. Every time a new neuron is identified as part of the neighborhood, it is added to $B$ and a recursive call is made with the radius decremented by one unit\footnote{This fact will ensure that the algorithm reaches the base case when $r=0$.}, marking the recently added neuron as the current node.

\begin{figure}[!ht]
  \centering
  \subfloat[Before. \label{fig:bubble1}]{
    \includegraphics[width=6.8cm]{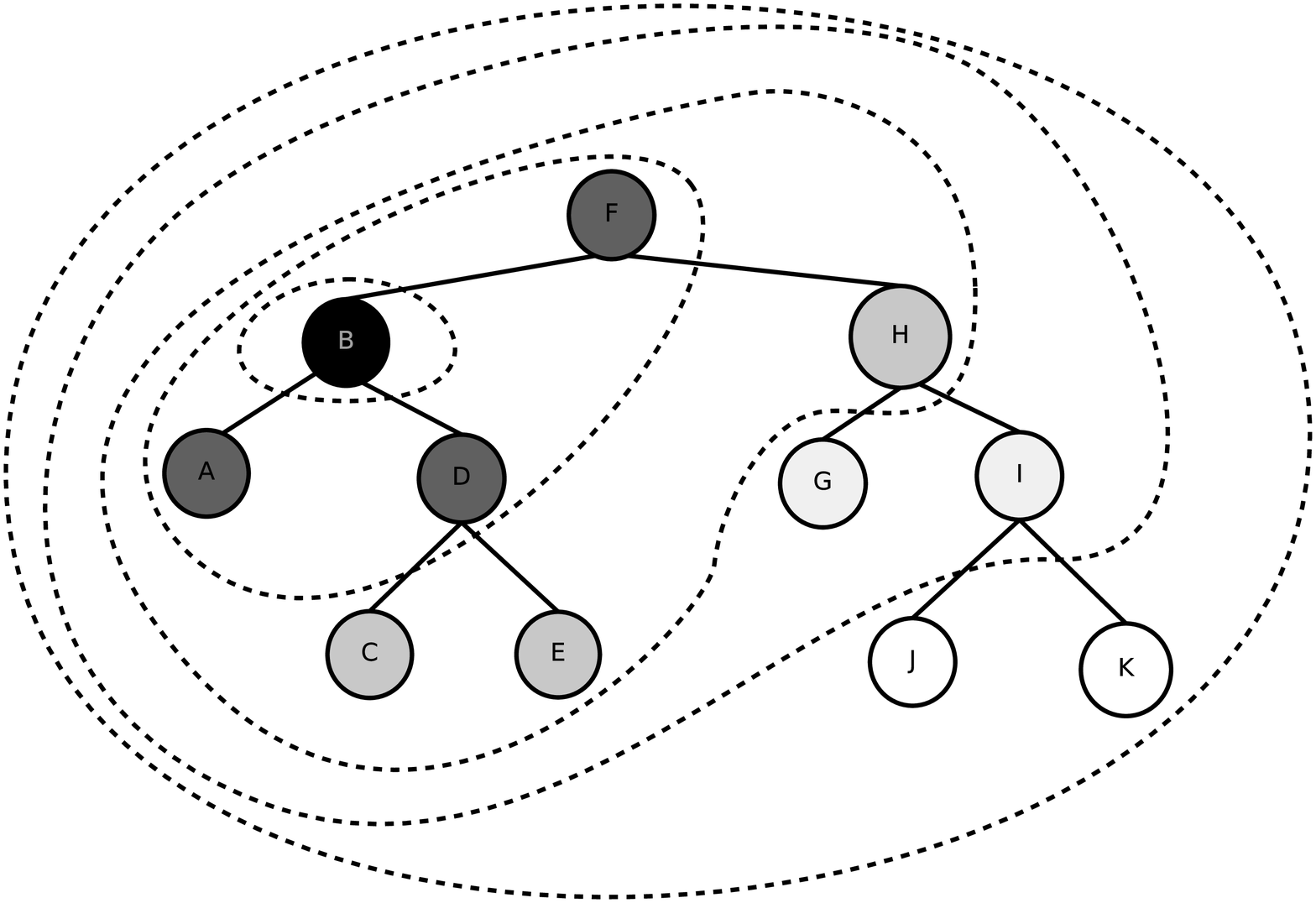}
  }
  \hspace{2mm}
  \subfloat[After. \label{fig:bubble2}]{
    \includegraphics[width=6.8cm]{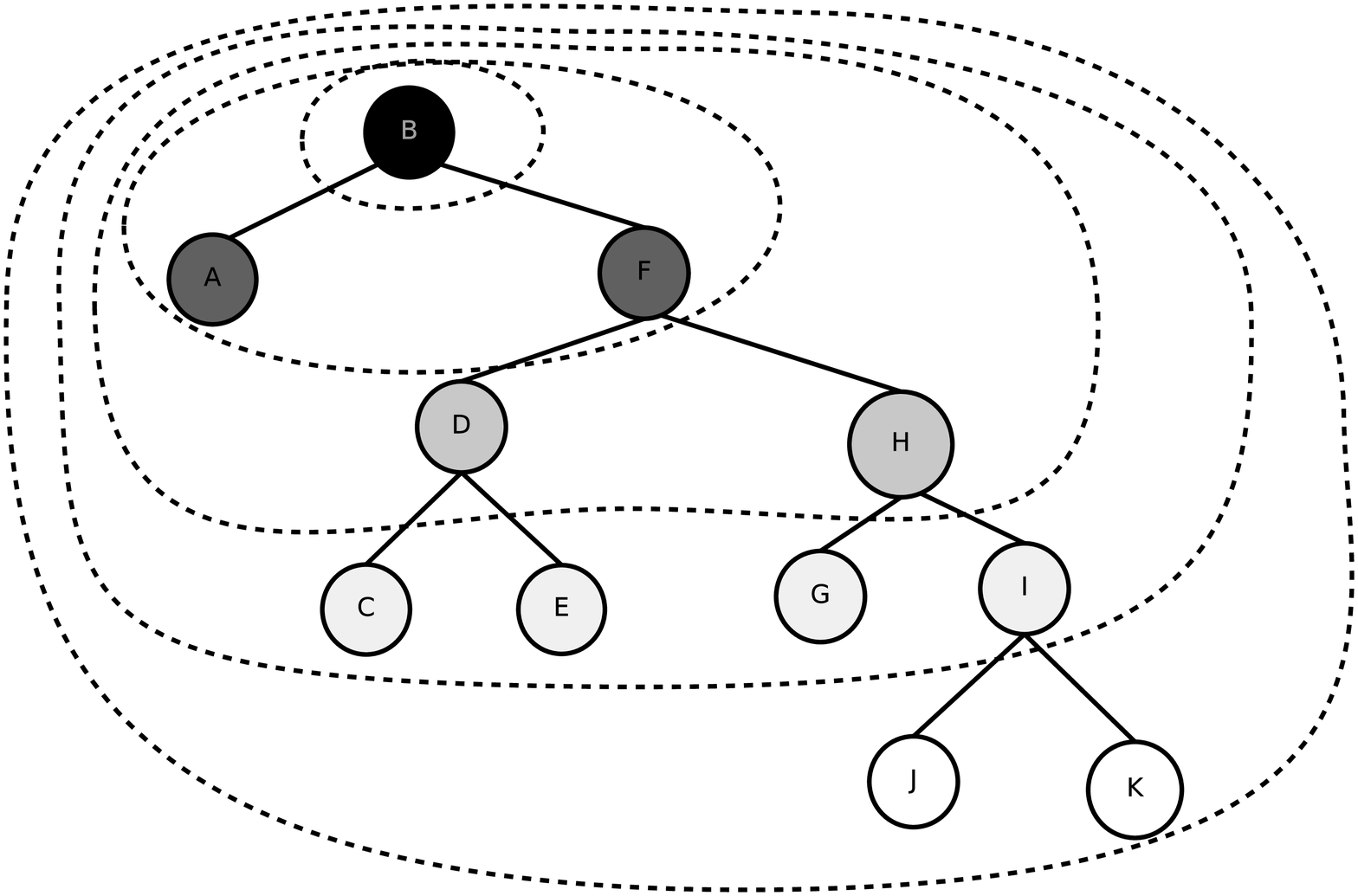}
  }
    \caption[The \ac{BoA} associated with the \ac{TTOSOM} before and after a rotation.]{The \ac{BoA} associated with the \ac{TTOSOM} before and after a rotation is invoked at node $B$.}
  \label{fig:TTOROT-bubble}
\end{figure}

The question of whether or not a neuron should be part of the current bubble, depends on the number of connections that separate the nodes rather than the distance that separate the networks in the solution space (for instance, the Euclidean distance). Figure \ref{fig:TTOROT-bubble} depicts how the \ac{BoA} differs from the one defined by the \ac{TTOSOM} as a result of applying a rotation. Figure \ref{fig:bubble1} shows  the \ac{BoA} around the node $B$, using the same configuration of the tree as in Figure \ref{fig:before-rotation}, i.e., before the rotation takes place. Here, the \ac{BoA} when $r=1$ involves the nodes $\{B,A,D,F\}$, and when $r=2$ the nodes contained in the bubble are $\{B,A,D,F,C,E,H\}$. Subsequently, considering a radius equal to $3$, the resulting \ac{BoA} contains the nodes $\{B,A,D,F,C,E,H,G,I\}$. Finally, the $r=5$ case leads to a \ac{BoA} which includes the whole set of nodes. Now, observe the case presented in Figure \ref{fig:bubble2}, which corresponds to the \ac{BoA} around $B$ \textit{after} the rotation upwards has been effected, i.e., the same configuration of the tree used in Figure \ref{fig:after-rotation}. In this case, when the radius is unity, nodes $\{B,A,F\}$ are the \textit{only} nodes within the bubble, which is different from the corresponding bubble before the rotation is invoked. Similarly, when $r=2$, we obtain a set different from the analogous pre-rotation case, which in this case is $\{B,A,F,D,H\}$. Note that coincidentally, for the case of a radius equal to 3, the bubbles are identical before and after the rotation, i.e., they invoke the nodes $\{B,A,D,F,G,I\}$. Trivially, again, when $r=5$, the \ac{BoA} invokes the entire tree.



As explained, Equation (\ref{eq:bubble-static}) describes the criteria for a \ac{BoA} calculated on a \textit{static} tree. It happens that, as a result of the conditional rotations, the tree will be dynamically adapted, and so the entire phenomenon has to be re-visited. Consequently, the \ac{BoA} around a particular node becomes a function of time, and, to reflect this fact, Equation (\ref{eq:bubble-static}) should be reformulated as:

\begin{equation}
 B(v_i;T,r,t)=\lbrace v | d_N(v_i,v;T,t) \leq r \rbrace,
 \label{eq:bubble-dynamic}
\end{equation}

\noindent where $t$ is the discrete time index.

The algorithm to obtain the \ac{BoA} for a specific node in such a setting is identical to Algorithm \ref{calculateneighborhood}, except that the input tree itself dynamically changes. Further, even though the formal notation includes the time parameter, ``$t$'', it happens that, in practice, the latter is needed only if the user/application requires a history of the \ac{BoA} for any or all the nodes. Storing the history of \acp{BoA} will require the maintenance of a \ac{DS} that will primarily store the changes made to the tree itself. Although storing the history of changes made to the tree can be done optimally \cite{Kaplan2004}, the question of explicitly storing the entire history of the \acp{BoA} for all the nodes in the tree remains open.


\subsubsection{Enforcing the \ac{BST} Property}

The \ac{CONROT-BST} heuristic \cite{Cheetham1993} requires that the tree should possess the \ac{BST} property \cite{Cormen2001}:

\textit{Let $x$ be a node in a \ac{BST}. If $y$ is a node in the left subtree of $x$, then $key[y] \leq key[x]$. Further, if $y$ is a node in the right subtree of $x$, then $key[x] \leq key[y]$.}

To satisfy the \ac{BST} property, first of all we see that, the tree must be binary\footnote{Of course, this is a severe constraint. But we are forced to require this, because the phenomenon of achieving conditional rotations for arbitrary $k$-ary trees is unsolved. This research, however, is currently being undertaken.}. As a general \ac{TTOSOM} utilizes an arbitrary number of children per node, one possibility is to bound the value of the branching factor to be $2$. In other words, the tree trained by the \ac{TTOSOM} is restricted to contain at most two children per node. Additionally, the tree must implicitly involve a comparison operator between the two children so as to discern between the branches and thus perform the search process. This comparison can be achieved by defining a unique key that must be maintained for each node in the tree, and which will, in turn, allow a lexicographical arrangement of the nodes.

This leads to a different, but closely related concept, which concerns the preservation of the topology of the \ac{SOM}. During the training process, the configuration of the tree will change as the tree evolves, positioning nodes that are accessed more often closer to the root. This probability-based ordering, will hopefully, be preserved by the rotations.

A particularly interesting case occurs when the imposed tree corresponds to a \textit{list} of neurons, i.e., a 1-ary tree. If the \ac{TTOSOM} is trained using such a tree where each node has at most two children, then the adaptive process will alter the original list. The rotations will then modify the original configuration, generating a new state, where the non-leaf nodes might have one or two children each. In this case the consequence of incorporating \ac{ADS}-based enhancements to the \ac{TTOSOM} will imply that the results obtained will be significantly different from those shown in \cite{Astudillo2011TTOSOM}.



As shown in \cite{Knuth1998}, an optimal arrangement of the nodes of the tree can be obtained using the probabilities of accesses. If these probabilities are not known \textit{a priori}, then the \ac{CONROT-BST} heuristic offers a solution, which involves a decision of whether or not to perform a single rotation towards the root. It happens that the concept of the ``just accessed'' node in the \ac{CONROT-BST} is compatible with the corresponding \ac{BMU} defined for the \ac{CL} model. In \ac{CL}, a neuron may be accessed more often than others and some techniques take advantage of this phenomenon through the inclusion of strategies that add or delete nodes

The \ac{CONROT-BST} implicitly stores the information acquired by the currently accessed node by incrementing a counter for that node. This is (in a distant sense) akin to the concept of a \ac{BMU} counter which adds or delete nodes in competitive networks.

During the training phase, when a neuron is a frequent winner of the \ac{CL}, it gains prominence in the sense that it can represent more points from the original data set. This phenomenon is registered by increasing the \ac{BMU} counter for that neuron.  We propose that during the training phase, we can verify if it is worth modifying the configuration of the tree by moving this neuron one level up towards the root as per the \ac{CONROT-BST} algorithm, and consequently explicitly recording the relevant role of the particular node with respect to its nearby neurons. \ac{CONROT-BST} achieves this by performing a \textit{local} movement of the node, where only its direct parent and children are aware of the neuron promotion.

\textbf{Neural Promotion} is the process by which a neuron is relocated in a more privileged position\footnote{As far as we know, we are not aware of any research which deals with the issue of Neural Promotion. Thus, we believe that this concept, itself, is pioneering.} in the network with respect to the other neurons in the \ac{NN}. Thus, while all ``all neurons are born equal'', their importance in the society of neurons is determined by what they represent. This is achieved, by an explicit  advancement of its rank or position. Given this premise, the nodes in the tree will be adapted in such a way that neurons that have been \acp{BMU} more frequently, will tend to move towards the root if an only if a reduction in the overall \ac{WPL} is obtained as a consequence of such a promotion. The properties of \ac{CONROT-BST} guarantee this.

\begin{figure}[!ht]
  \centering
  \includegraphics[scale=0.3]{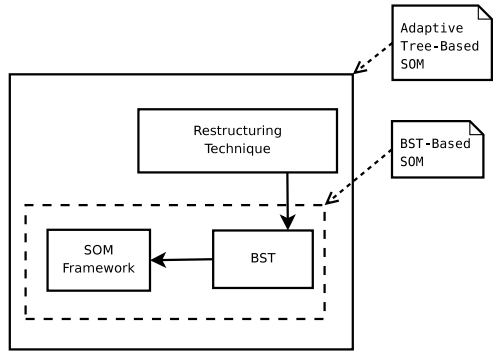}
  \caption{Architectural view of an Adaptive Tree-Based \ac{SOM}.}\label{fig:architecture}
\end{figure}

Once the \ac{SOM} and \ac{BST} are ``tied'' together in a symbiotic manner (where one enhances the other and vice versa), the adaptation can be achieved by affecting the configuration of the \ac{BST}. This task will be performed every time a training step of the \ac{SOM} is performed. Clearly, it is our task to achieve an integration of the \ac{BST} and the \ac{SOM}, and Figure \ref{fig:architecture} depicts the main architecture used to accomplish this. It transforms the structure of the \ac{SOM} by modifying the configuration of the \ac{BST} that, in turn, holds the structure of the neurons.

\CHECK
As this work constitutes the first attempt to constraint a tree-based \ac{SOM} using a \ac{BST}, our focus is placed on the self-adaptation of the nodes. In this sense, the unique identifiers of the nodes are employed to maintain the \ac{BST} structure and to promote nodes that are frequently accessed towards the root. We are currently examining ways to enhance this technique so as to improve the time required to identify the \ac{BMU} as well.
\NORMAL

\subsubsection{Initialization}

Initialization, in the case of the \ac{BST}-based \ac{TTOSOM}, is accomplished in two main steps which involve defining the initial value of each neuron and the connections among them. The initialization of the codebook vectors are performed in the same manner as in the basic \ac{TTOSOM}. The neurons can assume a starting value arbitrarily, for instance, by placing them on randomly selected input samples. On the other hand, a major enhancement with respect to the basic \ac{TTOSOM} lays in the way the neurons are linked together. The basic definition of the \ac{TTOSOM} utilizes connections that remain static through time. The beauty of such an arrangement is that it is capable of reflecting the user's perspective at the time of describing the topology, and it is able to preserve this configuration until the algorithm reaches convergence. The inclusion of the rotations renders this dynamic.

\subsubsection{The Required Local Information}

\begin{figure}[!ht]
  \centering
  \includegraphics[width=10cm]{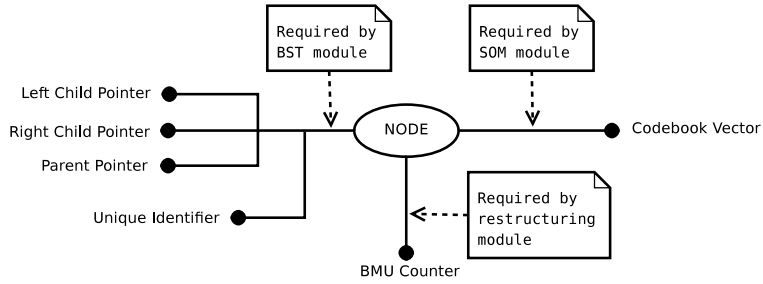}
  \caption{Fields included in a \ac{BST}-based \ac{SOM} neuron.}\label{fig:node}
\end{figure}

In our proposed approach, the codebooks of the \ac{SOM} correspond to the nodes of a \ac{BST}. Apart from the information regarding the codebooks themselves in the feature space, each neuron requires the maintenance of additional fields to achieve the adaptation. Besides this, each node inherits the properties of a \ac{BST} Node, and it thus includes a pointer to the left and right children, as well as (to make the implementation easier), a pointer to its parent. Each node also contains a label which is able to uniquely identify the neuron when it is in the ``company'' of other neurons. This identification index constitutes the lexicographical key used to sort the nodes of the tree and remains static as time proceeds. Figure \ref{fig:node} depicts all the fields included in a neuron of a \ac{BST}-based \ac{SOM}.

\subsubsection{The Neural State}

\begin{figure}[!ht]
  \centering
  \includegraphics[width=12.0cm]{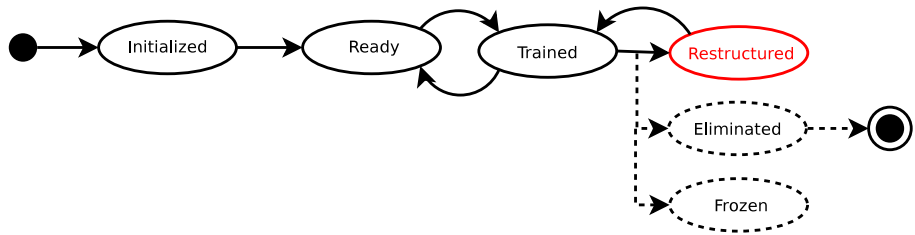}
  \caption{Possible states that a neuron may assume.}\label{fig:neural-state}
\end{figure}

The different states that a neuron may assume during its lifetime are illustrated in Figure \ref{fig:neural-state}. At first, when the node is created, it is assigned a unique identifier, and the rest of the data fields are populated with their initial values. Here, the codebook vector assumes a starting value in the feature space, and the pointers are configured so as to appropriately link the neuron with the rest of the neurons in the tree in a \ac{BST} configuration. Next, during the most significant portion of the algorithm, the \ac{NN} enters a main loop, where training is effected. This training phase, involves adjusting the codebooks and may also trigger optional modules that affect the neuron. Once the \ac{BMU} is identified, the neuron might assume the ``restructured'' state, which means that a restructuring technique, such as the \ac{CONROT} algorithm, will be applied. Alternatively, the neuron might be ready to accept queries, i.e., be part of the \ac{CL} process in the mapping mode.
Additionally, an option that we are currently investigating, involves the case when a neuron is no longer necessary and may thus be eliminated from the main neural structure. We refer to this state as the so-called ``deleted'' state, and it is depicted using dashed lines. Finally, we foresee an alternative state referred to as the ``frozen'' state, in which the neuron does not participate in the \ac{CL} during the training mode  although it may continue to be part of the overall \ac{NN} structure.

\subsubsection{The Training Step of the \ac{TTOCONROT} }

The training module of the \ac{TTOCONROT} is responsible of determining the \ac{BMU}, performing restructuring, calculating the \ac{BoA} and migrating the neurons within the \ac{BoA}.
Basically, what has to be done, is to integrate the \ac{CONROT} algorithm into the sequence of steps responsible for the training phase of the \ac{TTOSOM}. Algorithm \ref{alg:tto-conrot:training} describes the details of how this integration is accomplished. Line 1 performs the first task of the algorithm, which involves determining the \ac{BMU}. After that, line 2 invokes the \ac{CONROT} procedure. The rationale for following this sequence of steps is that the parameters needed to perform the conditional rotation, as specified in \cite{Cheetham1993}, includes the ``key'' of the element queried, which, in the present context, corresponds to the identity of the \ac{BMU}. At this stage of the algorithm, the \ac{BMU} may be rotated or not depending on the optimizing criterion given by equations (\ref{eq:psi-l}) and (\ref{eq:psi-r}), and the \ac{BoA} is determined \textit{after} this restructuring is done. These are performed in lines 3 and 4 of the algorithm respectively. Finally, lines 5 to 7, are responsible for the neural migration itself, and oversee the movement of the neurons within the \ac{BoA} towards the input sample.

\begin{algorithm}
\caption{\texttt{TTOCONROT-BST\_train}($x$,$p$)} \label{alg:tto-conrot:training}
\INPUT
\begin{description}
\item[i)]
 $x$, a sample signal.
 \vspace{-3mm}
\item[ii)]
 $p$, the pointer to the tree.
\end{description}
\METHOD
\begin{algorithmic} [1]
\STATE $v$ $\leftarrow$ \texttt{TTOSOM\_Find\_BMU}($x$,$p$)
\STATE \texttt{cond-rot-bst}($p$,$v.$\texttt{getID}())
\STATE $B$ $\leftarrow$ \{$v$\}
\STATE \texttt{TTOSOM\_Calculate\_Neighborhood}($B$,$v$,\textit{radius})
\FORALL {$b$ $\in$ $B$}
    \STATE \texttt{update\_rule}($b$.\texttt{getCodebook}(),$x$)
\ENDFOR
\end{algorithmic}
\ENDMETHOD
\end{algorithm}

\subsubsection{Alternative Restructuring Techniques}

Even though, we have explained the advantages of the \ac{CONROT} algorithm, the architecture that we are proposing allows the inclusion of  alternative restructuring modules other than the \ac{CONROT}. Potential candidates which can be used to perform the adaptation are the ones mentioned in Section \ref{sec:ads} and include the splay and the \ac{MT} algorithms, among others.

\section{Experimental Results}
\label{sec:exp}

\CHECK
To \textit{illustrate} the capabilities of our method, the experiments reported in the present work are primarily focused in lower dimensional feature spaces. This will help the reader in geometrically visualizing the results we have obtained.
\NORMAL
However, it is important to remark that the algorithm is also capable of solving problems in higher dimensions, although a graphical representation of the results will not be as illustrative. We know that, as per the results obtained in \cite{Astudillo2011TTOSOM}, the \ac{TTOSOM} is capable of inferring the distribution and structure of the data. However, in this present setting, we are interested in investigating the effects of applying the neural rotation as part of the training process. To render the results comparable, the experiments in this section use the same schedule for the learning rate and radius, i.e., no particular refinement of the parameters has been done for any specific data set. Additionally, the parameters follow a rather ``slow'' decrement of the so-called decay parameters, allowing us to understand how the prototype vectors are moved as convergence takes place. When solving practical problems, we recommend a further refinement of the parameters so as to increase the speed of the convergence process.

\subsection{\ac{TTOCONROT}'s Structure Learning Capabilities}

We shall describe the performance of \ac{TTOCONROT} with data sets in $1$, $2$ and $3$ dimensions, as well as experiments in the multidimensional domain. The specific advantages of the algorithm for various scenarios will also be highlighted.

\subsubsection{One Dimensional Objects}

\begin{figure}[!hbt]
  \centering
  \vspace{-2cm}
  \subfloat[After 0 iterations\label{fig:tto-conrot:curve:1}]{
    \includegraphics[width=7cm]{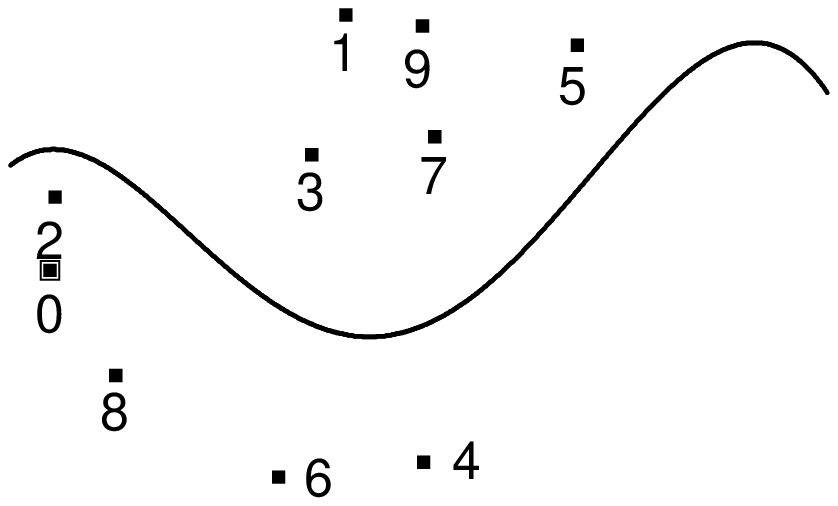}
  }
  \subfloat[After 1,000 iterations\label{fig:tto-conrot:curve:2}]{
    \includegraphics[width=7cm]{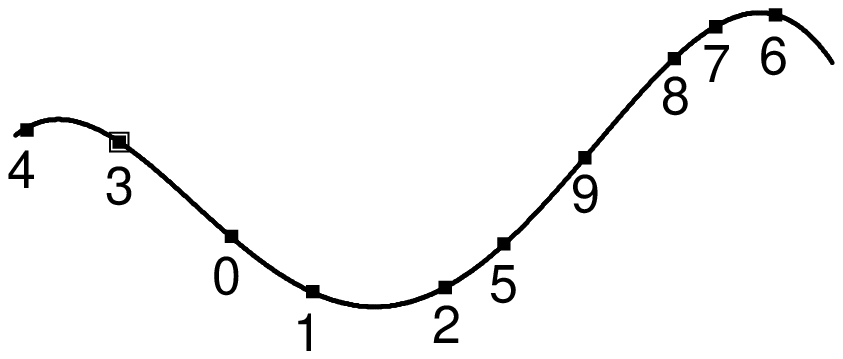}
  }
  \\
  \vspace{-2cm}
  \subfloat[After 3,000 iterations\label{fig:tto-conrot:curve:3}]{
    \includegraphics[width=7cm]{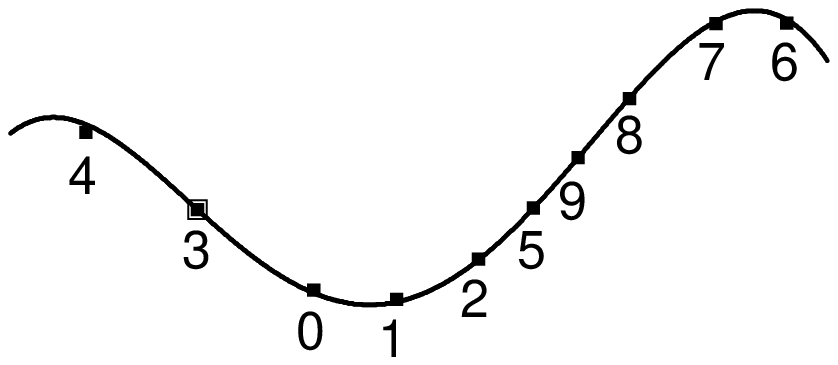}
  }
  \subfloat[After 5,000 iterations\label{fig:tto-conrot:curve:4}]{
    \includegraphics[width=7cm]{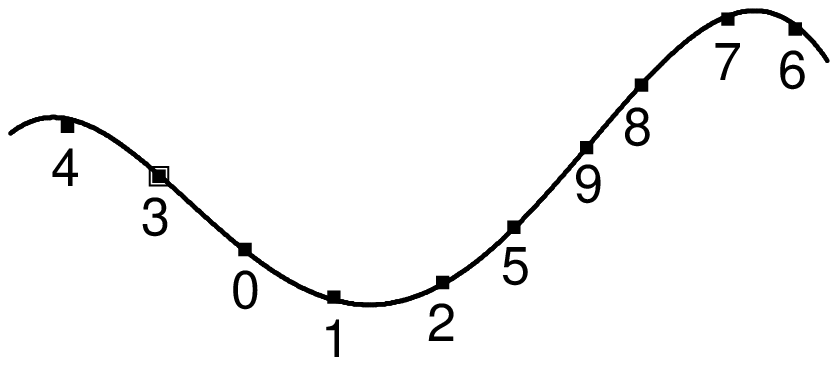}
  }
    \caption{ A 1-ary tree, i.e., a list topology, learns a curve. For the sake of simplicity, the edges are ommitted.}
  \label{fig:tto-conrot:curve}
\end{figure}

Since our entire learning paradigm assumes that the data has a tree-shaped model, our first attempt was to see how the philosophy is relevant to a unidimensional object (i.e., a curve), which really possesses a ``linear'' topology. Thus,  as a \textit{prima facie} case, we tested the strength of the \ac{TTOCONROT} to infer the properties of data sets generated from linear functions in the plane. Figure \ref{fig:tto-conrot:curve} shows different snapshots of how the \ac{TTOCONROT} learns the data generated from a curve. Random initialization was used by uniformly drawing points from the unit square. Observe that the original data points do not \textit{lie in the curve}. Our aim here was to show how our algorithm could learn the structure of the data using \textit{arbitrary} (initial and ``non-realistic'') values for the codebook vectors. Figures \ref{fig:tto-conrot:curve:2} and \ref{fig:tto-conrot:curve:3} depict the middle phase of the training process, where the edges connecting the neurons are omitted for simplicity. It is interesting to see how, after a few hundred training steps, the original chaotic placement of the neurons are rearranged so as to fall within the line described by the data points. The final configuration is shown in Figure \ref{fig:tto-conrot:curve:4}. The reader should observe that after convergence has been achieved, the neurons are placed almost equidistantly along the curve. Even though the codebooks are not sorted in and increasing numerical order, the hidden tree and its root, denoted by two concentric squares, are configured in such a way that nodes that are queried more frequently will tend to be closer to the root. In this sense, the algorithm is not only capturing the essence of the topological properties of the data set, but at the same time rearranging the internal order of the neurons according to their importance in terms of their probabilities of access.

\subsubsection{Two Dimensional Data Points}

\begin{figure}[!hbt]
  \centering
  \subfloat[After 0 iterations\label{fig:tto-conrot:triangle:1}]{
    \includegraphics[width=7cm]{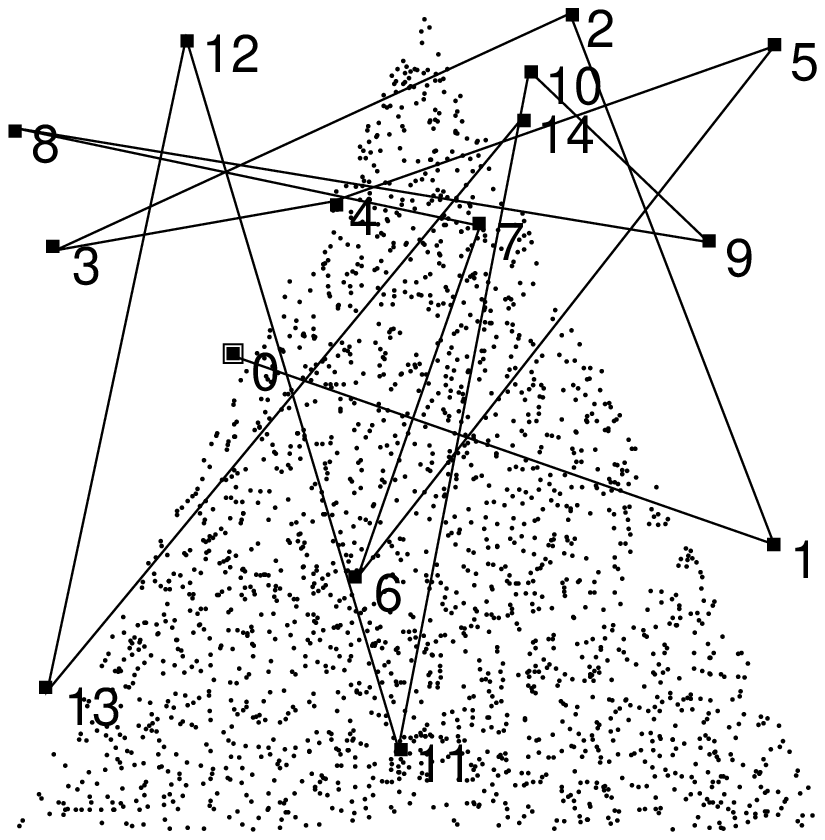}
  }
  \subfloat[After 1,000 iterations\label{fig:tto-conrot:triangle:2}]{
    \includegraphics[width=7cm]{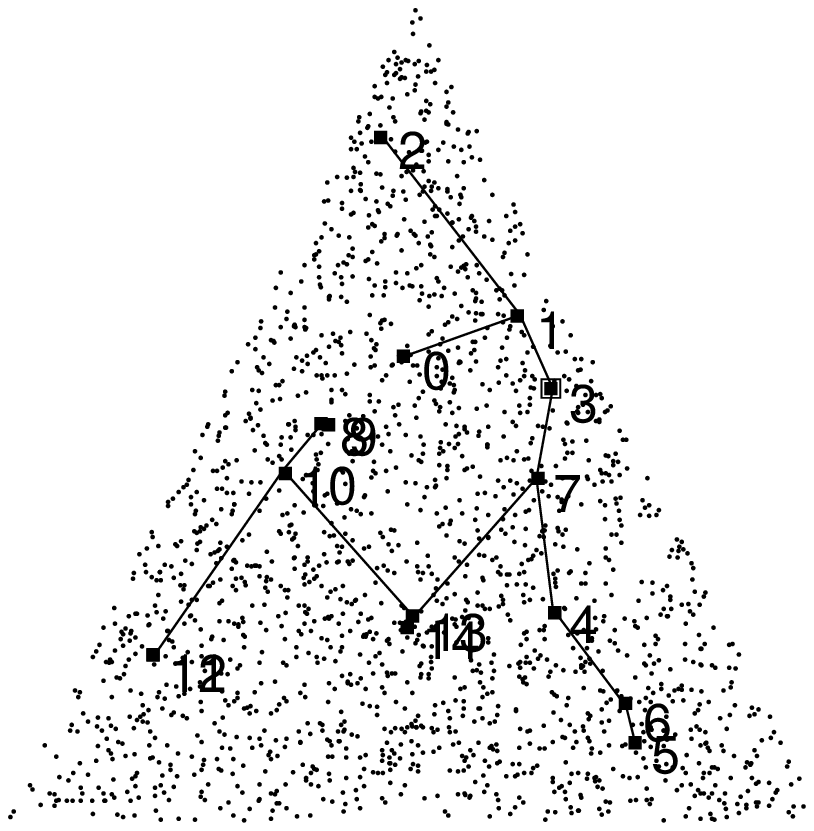}
  }
  \\
  \subfloat[After 3,000 iterations\label{fig:tto-conrot:triangle:3}]{
    \includegraphics[width=7cm]{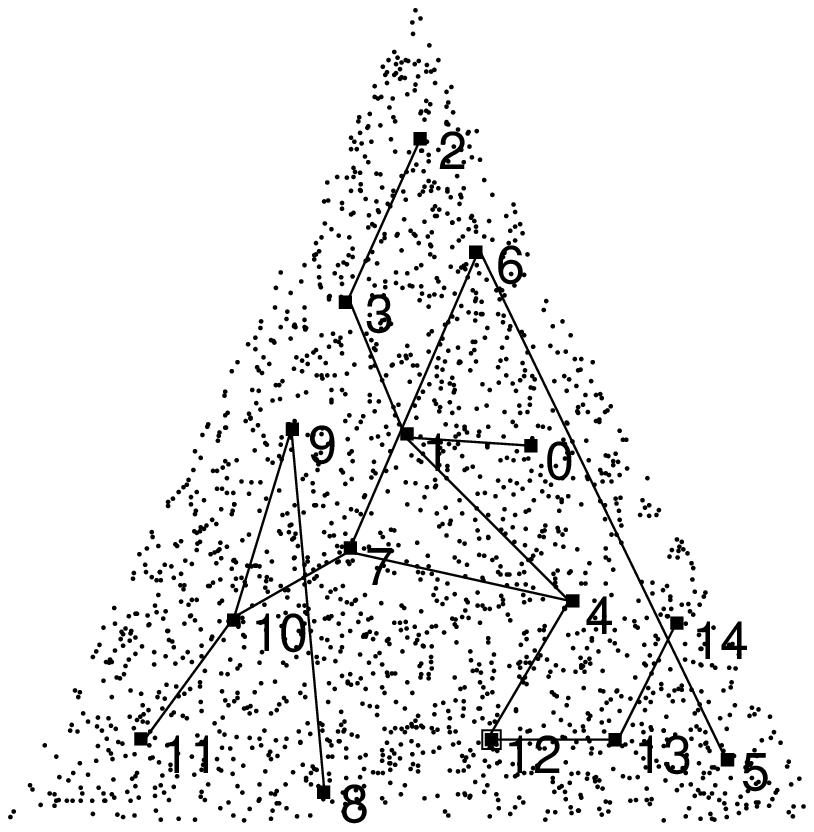}
  }
  \subfloat[After 5,000 iterations\label{fig:tto-conrot:triangle:4}]{
    \includegraphics[width=7cm]{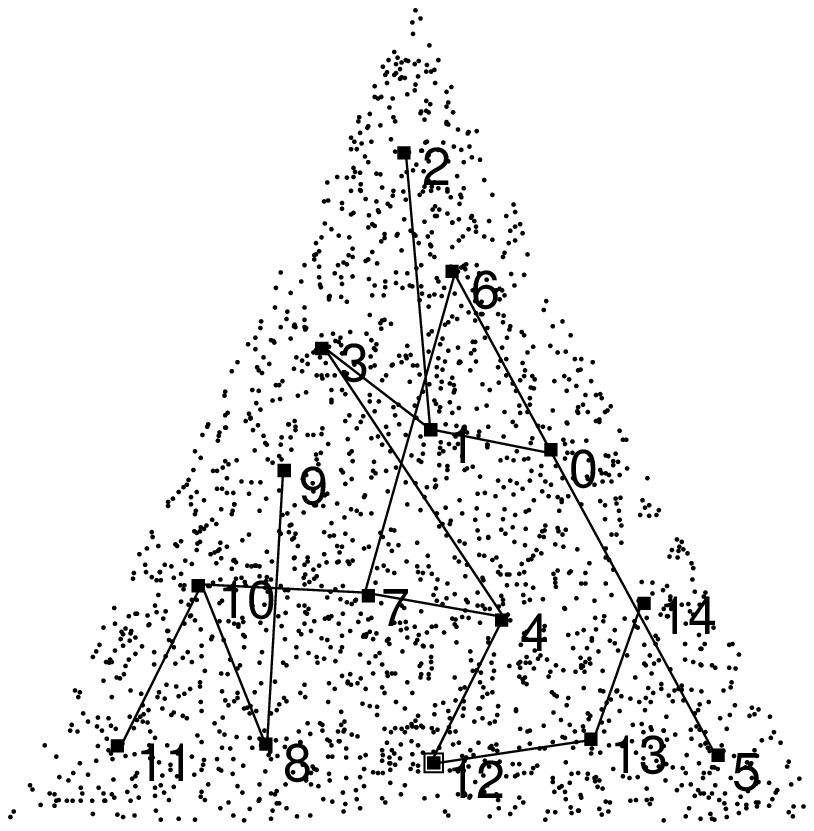}
  }
  \caption{  A 1-ary tree, i.e., a \textit{list} topology, learns a triangular distribution. The \ac{DS} is self-adapted so that nodes accessed more frequently are moved closer to the root conditionally. The \ac{BST} property is also preserved.}
  \label{fig:tto-conrot:triangle}
\end{figure}

To demonstrate the power of including \ac{ADS} in \acp{SOM}, we shall now consider the same two-dimensional data sets studied in \cite{Astudillo2011TTOSOM}. First we consider the data generated from a triangular-spaced distribution, as shown in Figures \ref{fig:tto-conrot:triangle:1}-\ref{fig:tto-conrot:triangle:4}. In this case,  the initial tree topology is unidirectional, i.e., a list, although, realistically, this is quite inadvisable considering the true (unknown) topology of the distribution. In other words, we assume that the user has \textit{no a priori} information about the data distribution. Thus, for the initialization phase, a 1-ary tree is employed as the tree structure, and the respective keys are assigned in an increasing order. Observe that in this way we are providing minimal information to the algorithm. The root of the tree is marked with two concentric squares, i.e., the neuron labeled with the index $0$ in Figure \ref{fig:tto-conrot:triangle:1}. Also, with regards to the feature space, the prototype vectors are initially randomly placed. In the first iteration, the linear topology is lost, which is attributable to the randomness of the data points. As the prototypes are migrated and reallocated (see Figures \ref{fig:tto-conrot:triangle:2} and \ref{fig:tto-conrot:triangle:3} ), the 1-ary tree is modified as a consequence of the \textit{rotations}. Such a transformation is completely novel to the field of \acp{SOM}. Finally, Figure \ref{fig:tto-conrot:triangle:4} depicts the case after convergence has taken place. Here, the tree nodes are uniformly distributed over the entire triangular domain. The \ac{BST} property is still preserved, and further rotations are still possible if the training process continues.

This experiment serves as an excellent example to show the differences between our current method and the original \ac{TTOSOM} algorithm \cite{Astudillo2011TTOSOM}, where the same data set with similar settings was utilized. In the case of the \ac{TTOCONROT} the points effectively represent the entire data set. However, the reader must observe that we do not have to provide the algorithm with any particular \textit{a priori} information about the structure of the data distribution -- this is learned during the training process, as shown in Figure \ref{fig:tto-conrot:triangle:4}. Thus, the specification of the initial ``user-defined'' tree topology (representing his perspective of the data space) required by the \ac{TTOSOM} is no longer mandatory, and an alternative specification which \textit{only} requires the number of nodes in the initial 1-ary tree is sufficient.

\begin{figure}[!hbt]
  \centering
    \includegraphics[width=7cm]{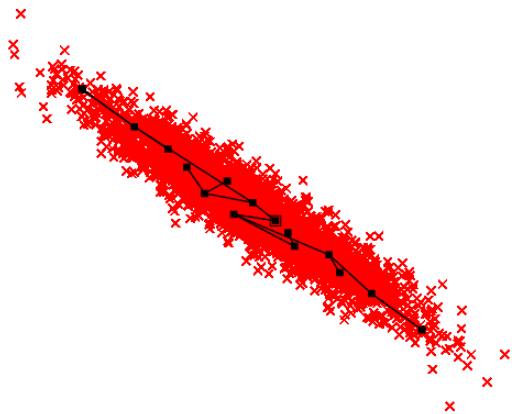}
  \caption{ A tree learns a Gaussian distribution. The neurons that are accessed more frequently are promoted closer to the root.}
  \label{fig:tto-conrot:gaussian15}
\end{figure}

A second experiment involves a Gaussian distribution. Here a 2-dimensional Gaussian ellipsoid is learned using the \ac{TTOCONROT} algorithm. The convergence of the entire training execution phase is displayed in Figure \ref{fig:tto-conrot:gaussian15}. This experiment considers a complete \ac{BST} of depth 4, i.e., containing 15 nodes. For simplicity the labels of the nodes have been removed.


In Figure \ref{fig:tto-conrot:gaussian15}, the tree structure generated by the neurons suggest an ellipsoidal structure for the data distribution. This experiment is a good example to show how the nodes close to the root represent dense areas of the ellipsoid, and at the same time, those node that are far from the root (in tree space) occupy regions with low density, e.g., in the ``extremes'' of the ellipse. The \ac{TTOCONROT} infers this structure without receiving any \textit{a priori} information about the distribution or its structure.

\begin{figure}[!ht]
  \centering
  \subfloat[After 0 iterations\label{fig:irregular-1}]{
    \includegraphics[width=3.5cm]{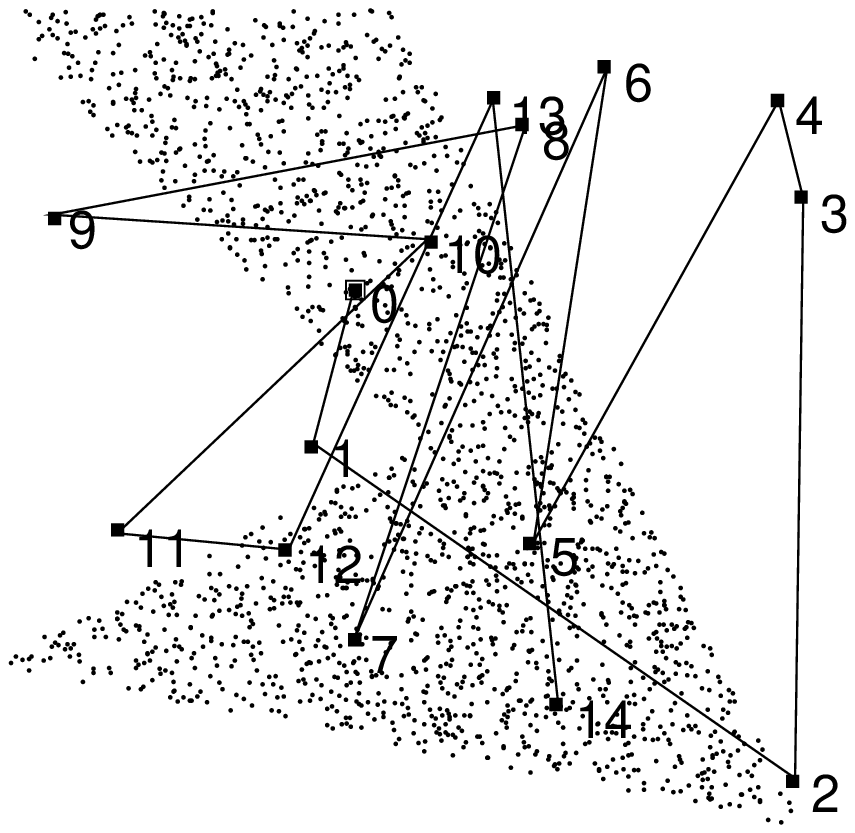}
  }
  \subfloat[After 1,000 iterations\label{fig:irregular-2}]{
    \includegraphics[width=3.5cm]{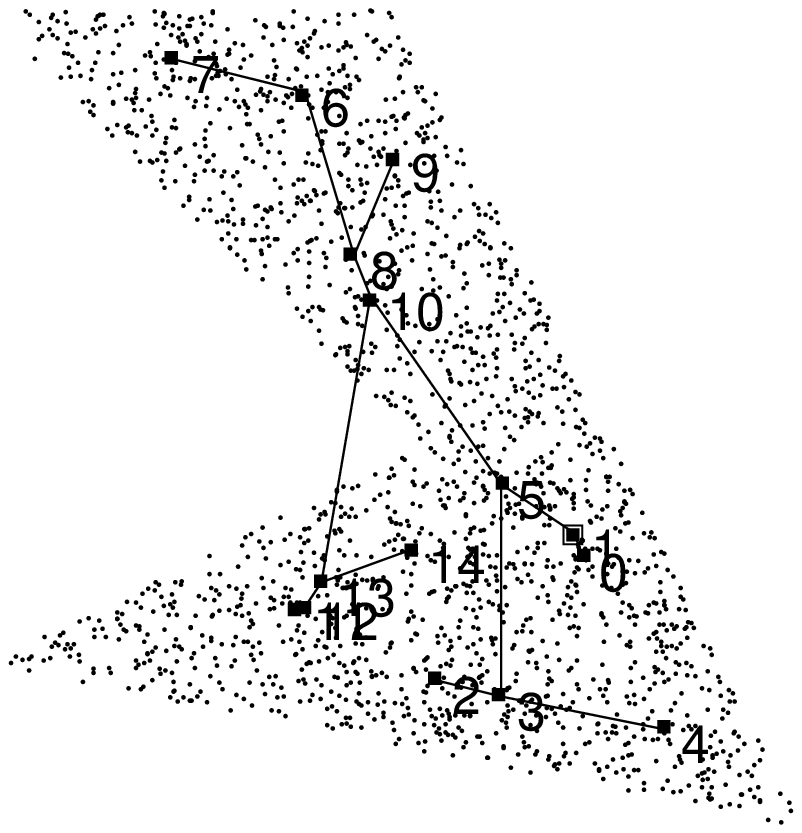}
  }
  \subfloat[After 3,000 iterations\label{fig:irregular-3}]{
    \includegraphics[width=3.5cm]{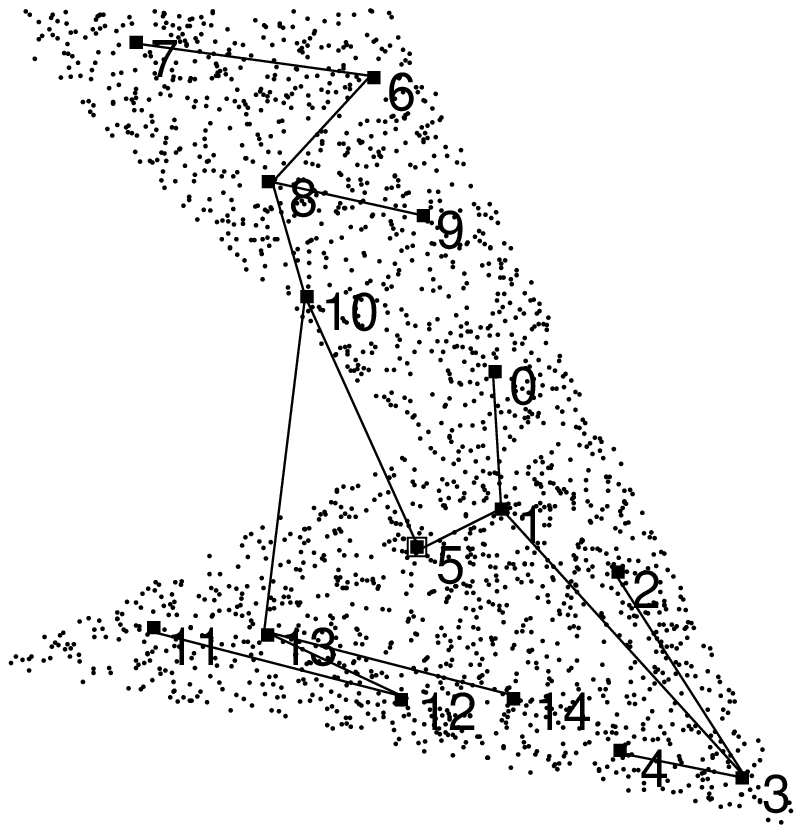}
  }
  \subfloat[After 5,000 iterations\label{fig:irregular-4}]{
    \includegraphics[width=3.5cm]{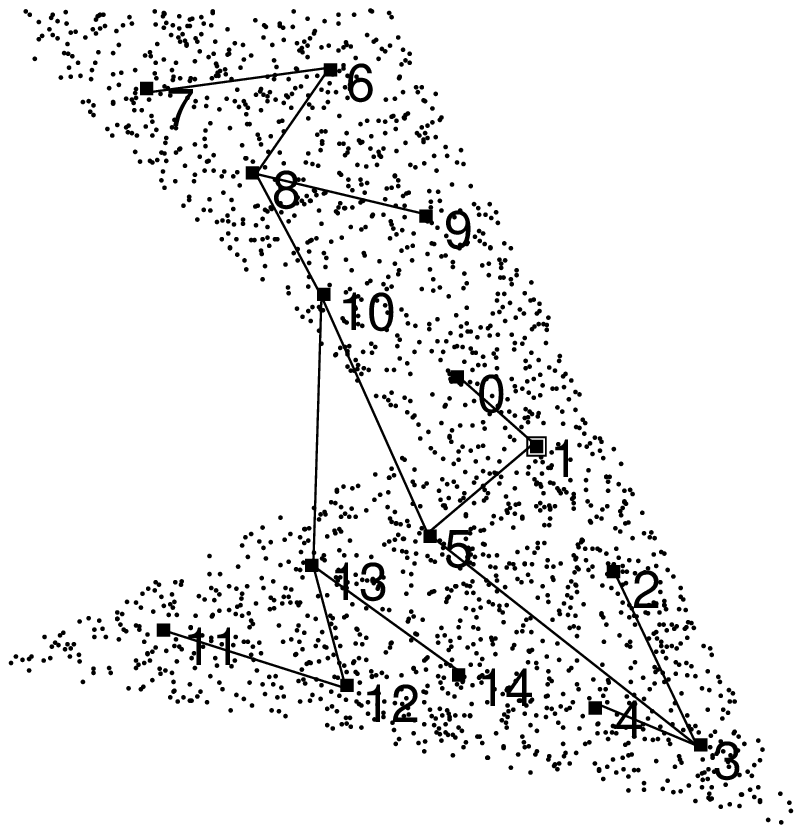}
  }
    \caption[A 1-ary tree learns different distributions from a concave object using the \ac{TTOCONROT} algorithm.]{ A 1-ary tree, i.e., a list topology, learns different distributions from a concave object using the \ac{TTOCONROT} algorithm. The set of parameters is the same as in the other examples.}
  \label{fig:tto-conrot:irregular}
\end{figure}

The experiment shown in Figures \ref{fig:irregular-1}-\ref{fig:irregular-4} considers data generated from an irregular shape with a concave surface. Again, as in the case of the experiments described earlier, the original tree includes 15 neurons arranged unidirectionally, i.e., as in a list. As a result of the training, the distribution is learned and the tree is adapted accordingly, as illustrated in Figure \ref{fig:irregular-4}. Observe that the random initialization is performed by randomly selecting points from the unit square, and this points thus do not necessarily fall within the concave boundaries. Although this initialization scheme is responsible of placing codebook vectors outside of the irregular shape, the reader should observe that in a few training steps, they are repositioned inside the contour.
\CHECK
It is important to indicate that, even though after the convergence of the algorithm, a line connecting two points passes outside the overall ``unknown'' shape, one must take into account that the \ac{TTOCONROT} tree attempts to mimic the stochastic properties in terms of access probabilities. When the user desires the topological mimicry in terms of skeletal structure, we recommend the use of the \ac{TTOSOM} instead.
\NORMAL
The final distribution of the points is quite amazing!

\subsubsection{Three Dimensional Data Points}

\begin{figure}[!hbt]
  \centering
  \subfloat[After 0 iterations\label{fig:tto-conrot:sphere:without-conrot:1}]{
    \includegraphics[width=6.2cm]{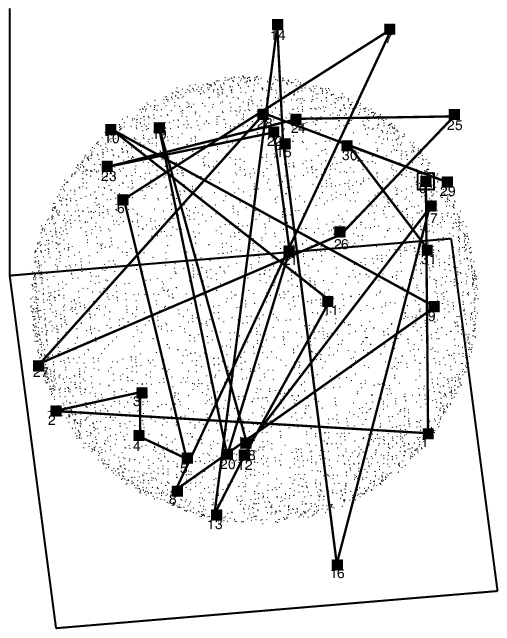}
  }
  \hspace{-35mm}
  \subfloat[After 1,000 iter.\label{fig:tto-conrot:sphere:without-conrot:2}]{
    \includegraphics[width=6.2cm]{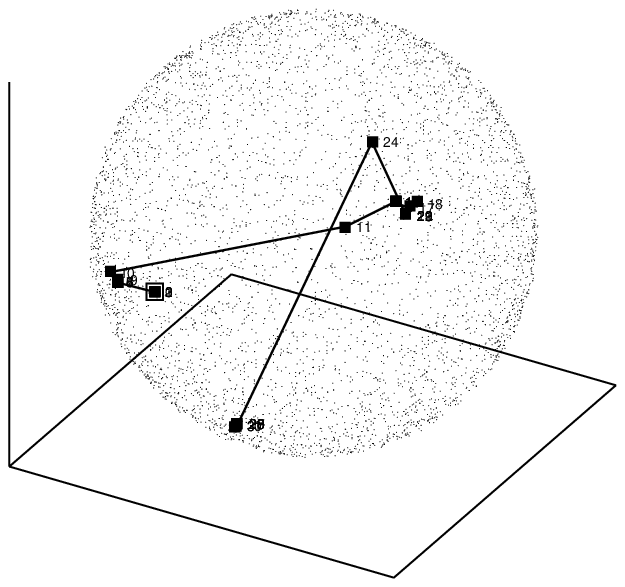}
  }
  \hspace{-35mm}
  \subfloat[After 3,000 iter.\label{fig:tto-conrot:sphere:without-conrot:3}]{
    \includegraphics[width=6.2cm]{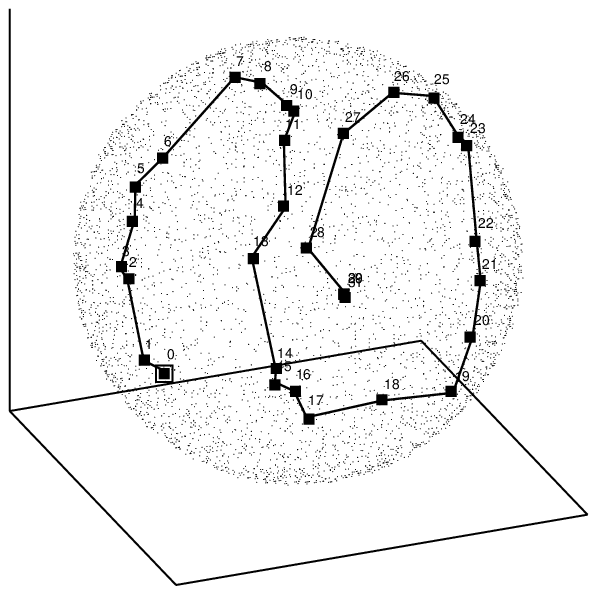}
  }
  \hspace{-35mm}
  \subfloat[After 5,000 iter.\label{fig:tto-conrot:sphere:without-conrot:4}]{
    \includegraphics[width=6.2cm]{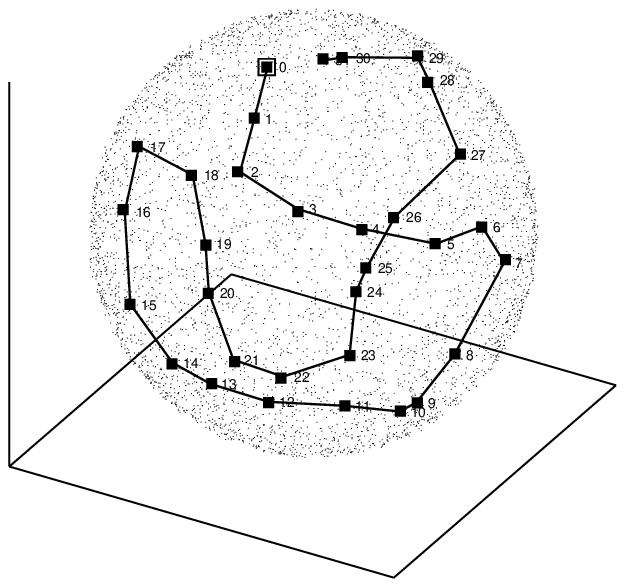}
  }
    \caption[A 1-ary tree learns a sphere distribution without conditional rotations.]{A 1-ary tree, i.e., a list topology, learns a sphere distribution when the algorithm does not utilize any conditional rotation.}
  \label{fig:tto-conrot:sphere:without-conrot}
\end{figure}

We will now explain the results obtained when applying the algorithm with and without \ac{CONROT}. To do this we opt to consider three-dimensional objects. The experiments utilize the data generated from the contour of the unit sphere. It also initially involves an \textit{uni}-dimensional chain of 31 neurons. Additionally, in order to show the power of the algorithm, both cases initialize the codebooks by randomly drawing points from the unit cube, which thus initially places the points outside the sphere itself. Figure \ref{fig:tto-conrot:sphere:without-conrot} presents the case when the basic TTO algorithm (without \ac{CONROT}) learns the unit sphere without performing conditional rotations. The illustration presented in Figure \ref{fig:tto-conrot:sphere:without-conrot:1} show the state of the neurons before the first iteration is completed. Here, as shown, the codebooks lie inside the unit cube, although some of the neurons are positioned outside the boundary of the respective circumscribed sphere, which is the one we want to learn. Secondly, Figures \ref{fig:tto-conrot:sphere:without-conrot:2} and \ref{fig:tto-conrot:sphere:without-conrot:3} depict intermediate steps of the learning phase. As the algorithm processes the information provided by the sample points and the neurons are repositioned, the chain of neurons is constantly ``twisted'' so as to adequately represent the entire manifold. Finally, Figure \ref{fig:tto-conrot:sphere:without-conrot:4} illustrates the case when the convergence is reached. In this case, the one-dimensional list of neurons is evenly distributed over the sphere, preserving the original properties of the 3-dimensional object and also presenting a shape which reminds the viewer of the so-called Peano curve \cite{Peano1890}.

\begin{figure}[!hbt]
  \centering
  \subfloat[After 0 iterations\label{fig:tto-conrot:sphere:1}]{
    \includegraphics[width=6.2cm]{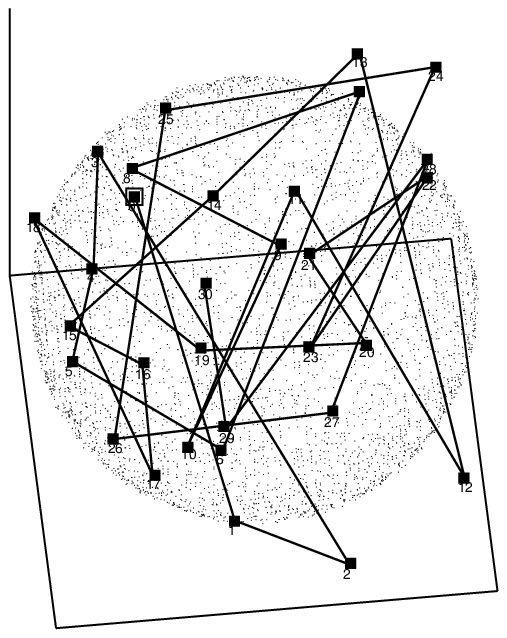}
  }
   \hspace{-35mm}
  \subfloat[After 1,000 iter.\label{fig:tto-conrot:sphere:2}]{
    \includegraphics[width=6.2cm]{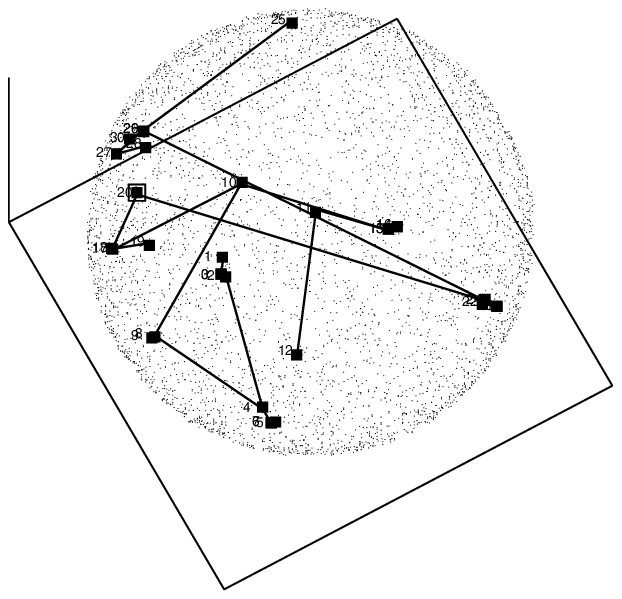}
  }
   \hspace{-35mm}
  \subfloat[After 3,000 iter.\label{fig:tto-conrot:sphere:3}]{
    \includegraphics[width=6.2cm]{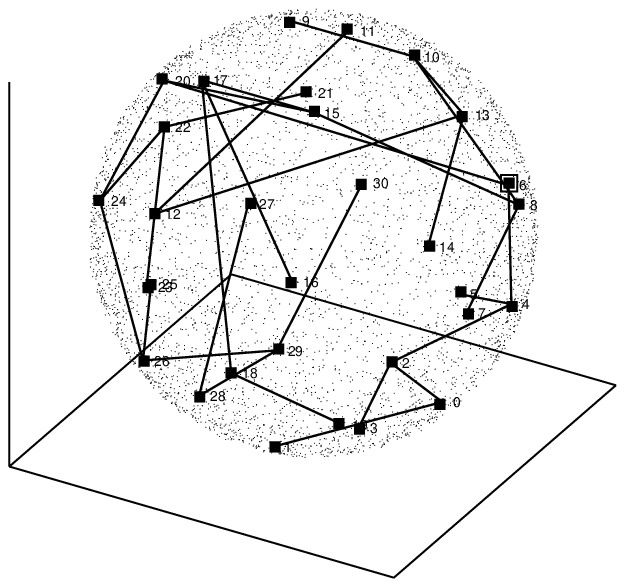}
  }
  \hspace{-35mm}
  \subfloat[After 5,000 iter.\label{fig:tto-conrot:sphere:4}]{
    \includegraphics[width=6.2cm]{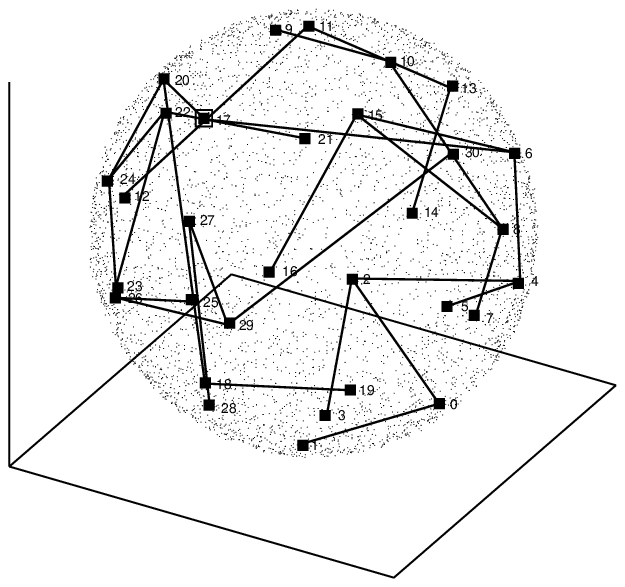}
  }
    \caption{  A 1-ary tree, i.e., a list topology, learns a sphere distribution.}
  \label{fig:tto-conrot:sphere}
\end{figure}

A complimentary set of experiments which involved the learning of the same unit sphere where the TTO scheme was augmented by conditional rotations (i.e., \ac{CONROT}) was also conducted.
Figure \ref{fig:tto-conrot:sphere:1} shows the initialization of the codebooks. Here, the starting positions of the neurons fall within the unit cube as in the case displayed in Figure \ref{fig:tto-conrot:sphere:without-conrot:1}. Figures  \ref{fig:tto-conrot:sphere:2} and \ref{fig:tto-conrot:sphere:3} show snapshots after $1,000$ and $3,000$ iterations respectively. In this case the tree configuration obtained in the intermediate phases differ significantly from those obtained by the corresponding configurations shown in Figure  \ref{fig:tto-conrot:sphere:without-conrot}, i.e., those that involved no rotations. In this case, the list rearranges itself as per \ac{CONROT}, modifying the original chain structure to yield a more-or-less balanced tree. Finally, from the results obtained after convergence, and illustrated in Figure \ref{fig:tto-conrot:sphere:4}, it is possible to compare both scenarios. In both cases, we see that the tree is accurately learned. However, in the first case, the structure of the nodes is maintained as a list throughout the learning phase, while, in the case when \ac{CONROT} is applied, the configuration of the tree is constantly revised, promoting those neurons that are queried more frequently. Additionally, the experiments show us how the dimensionality reduction property evidenced in the traditional \ac{SOM}, is also present in the \ac{TTOCONROT}. Here, an object in the 3-dimensional domain is successfully learned by our algorithm, and the properties of the original manifold are captured from the perspective of a tree.

\subsubsection{Multidimensional Data Points}

The well known Iris dataset was chosen for showing the power of our scheme in a scenario when the dimensionality is increased. This data set gives the measurements (in centimeters) of the variables which are the sepal length, sepal width, petal length and petal width, respectively, for 50 flowers from each of 3 species of the iris family. The species are the Iris Setosa, Versicolor, and Virginica.

\begin{figure}[!ht]
   \centering
   \subfloat[Using 7 nodes.\label{fig:conrot:iris14-7}]{
     \includegraphics[width=4.5cm]{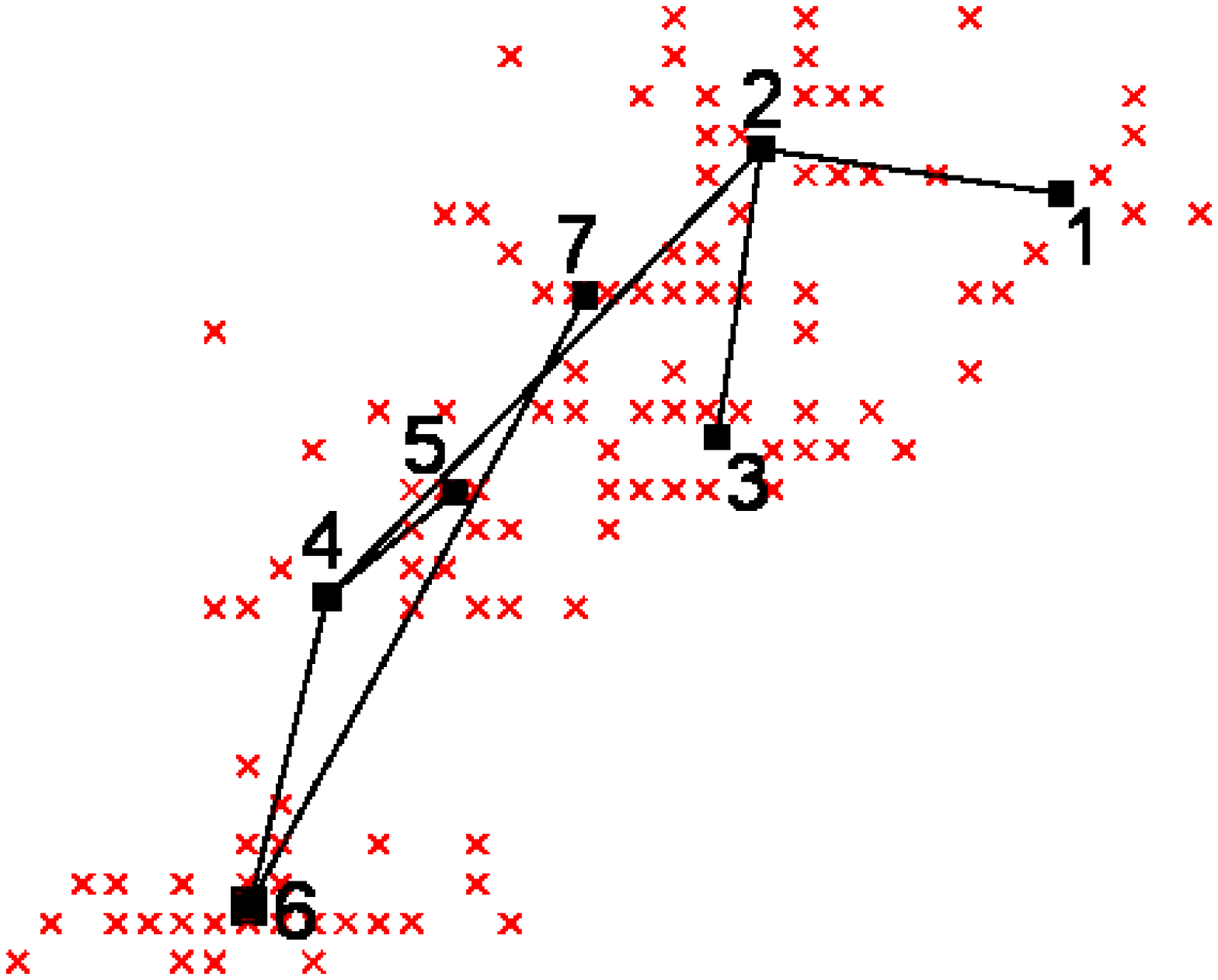}
   }
   \hspace{5mm}
   \subfloat[Using 15 nodes.\label{fig:conrot:iris14-15}]{
     \includegraphics[width=4.5cm]{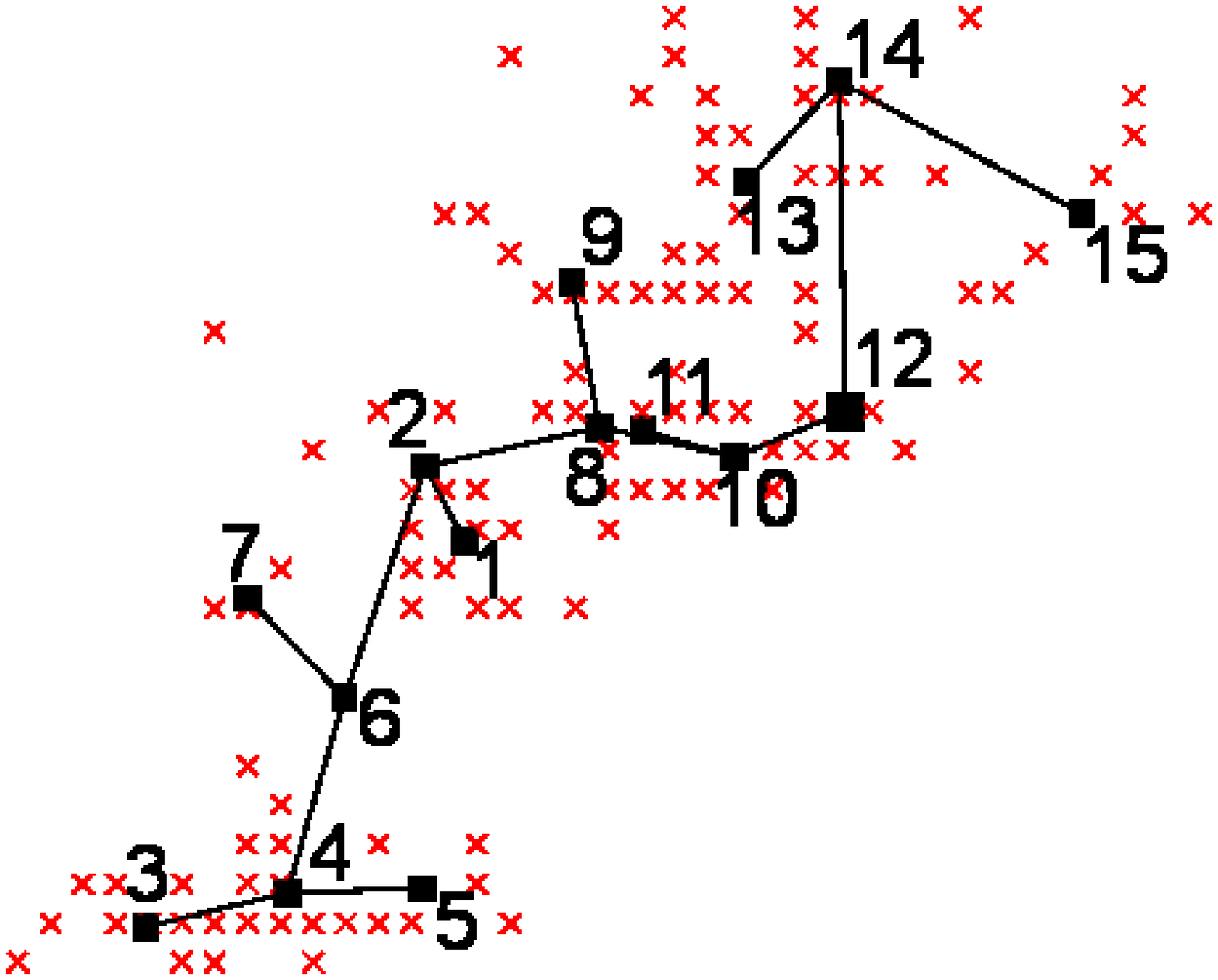}
   }
   \hspace{5mm}
      \subfloat[Using 31 nodes.\label{fig:conrot:iris14-31}]{
     \includegraphics[width=4.5cm]{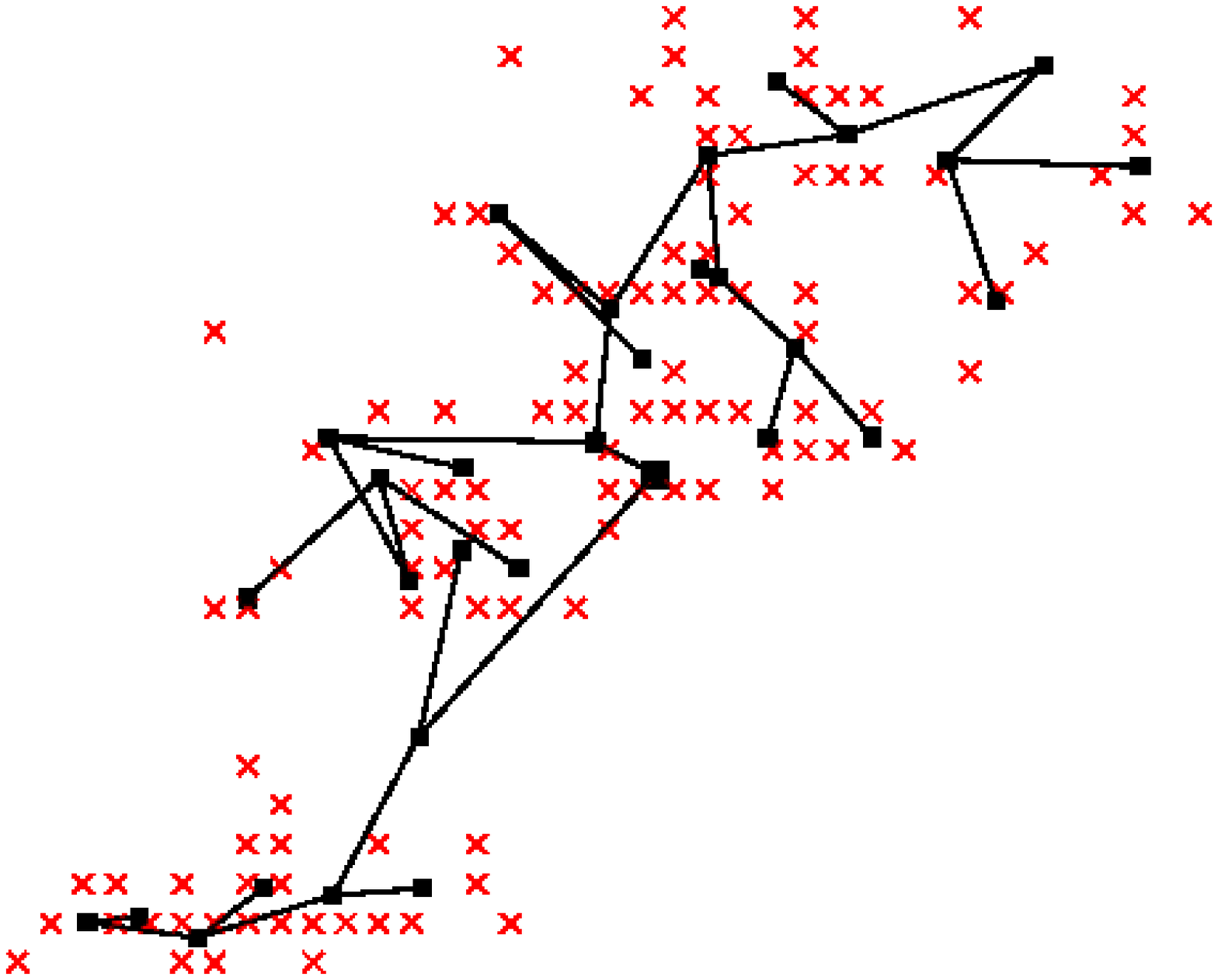}
   }
     \caption{Three different experiments where the \ac{TTOCONROT} effectively captures the fundamental structure of the iris dataset (2-dimensional projection of the data is shown).}
   \label{fig:conrot:iris14}
\end{figure}

In this set of experiments, the Iris data set was learned under three different configurations, using a fixed schedule for the learning rate and radius but with a distinct tree configuration. The results of the experiments are depicted in Figure \ref{fig:conrot:iris14} and involve a complete binary tree of depth 3, 4 and 5, respectively. Taking into account that the dataset possesses a high dimensionality, we present the projection in the 2-dimensional space to facilitate the visualization. We also removed the labels from the nodes in Figure \ref{fig:conrot:iris14-31} to improve understandability.

The experiment utilizes a underlying tree topology of a complete binary tree with different levels of depth. By this we attempt to show examples of how exactly the \textit{same} parameters of the TTOCONROT, can be utilized to learn the structure from data belonging to the 2-dimensional, 3-dimensional and also 4-dimensional spaces. After executing the TTO-SOM, each of the main branches of the tree were migrated towards the center of mass of the cloud of points in the hyper-space belonging to each of the three categories of flowers, respectively.

%
Since the TTOCONROT is an unsupervised learning algorithm, it performs learning without knowing the true labels of the samples. However, when these labels are available, one can use them to evaluate the quality of the tree. To do so, each sample is assigned to its closest neuron, and tagging the neuron with the class which is most frequent. Table \ref{table:iris14-7} presents the evaluation for the tree in Figure \ref{fig:conrot:iris14-7}.

\begin{table}[!htb]
\centering
\begin{tabular}{l*{7}{c}}
Assigned to neuron	$\rightarrow$	&  \textbf{1} &  \textbf{2}  &  \textbf{3}  & \textbf{4} &  \textbf{5}  &  \textbf{6}  &  \textbf{7} \\
\hline
\texttt{Iris-setosa}					&  0 &  0  &  0  & 0 &  0  & 50  &  0 \\
\texttt{Iris-versicolor}				&  0 &  0  & 20  & 7 & 20  &  0  &  3 \\
\texttt{Iris-virginica}				& 12 & 22  &  0  & 0 &  1  &  0  & 15 \\
\end{tabular}
\caption{``Cluster to class'' evaluation for the tree in Figure \ref{fig:conrot:iris14-7}.}
\label{table:iris14-7}
\end{table}

Using the simple voting scheme explained above, it is possible to see from Table \ref{table:iris14-7}, that only 4 instances are incorrectly classified, i.e., 97.3\% of the instances are correctly classified. Additionally, observe that node 6 contains all the 50 instances corresponding to the class \texttt{Iris-setosa}. It is well known that the  Iris-setosa class is linearly separable from the other two classes, and our algorithm was able to discover this without providing it with the labels. We find this result quite fascinating!

The experimental results shown in Table \ref{table:iris14-7}, not only demonstrate the potential capabilities of the \ac{TTOCONROT} for performing clustering, but also suggest the possibilities of using it for pattern classification. According to \cite{Duda2000}, there are several reasons for performing pattern classification using an unsupervised approach. We are currently investigating such a classification strategy.

\subsection{Skeletonization}

In general, the main objective of skeletonization consists of generating a simpler representation of the shape of an object. The authors of \cite{Ogniewicz1995} refer to skeletonization in the plane as the process by which a 2-dimensional shape is transformed into a 1-dimensional one, similar to a ``stick'' figure. The applications of skeletonization are diverse, including the fields of \acl{CV} and \acl{PR}.

As explained in \cite{Astudillo2011TTOSOM}, the traditional methods for skeletonization assume the connectivity of the data points and when this is not the case, more sophisticated methods are required. Previous efforts involving \ac{SOM} variants to achieve skeletonization has been proposed \cite{Astudillo2011TTOSOM,Datta1996,Singh2000}. We remark that the \ac{TTOSOM} \cite{Astudillo2011TTOSOM} is the only one which uses a tree-based structure. The \ac{TTOSOM} assumed that the shape of the object is not known \textit{a priori}. Rather, this is learned by accessing a single point of the entire shape at any time instant. Our results reported in \cite{Astudillo2011TTOSOM} confirm that this is actually possible, and we now focus on how the conditional rotations will affect such a skeletonization.

\begin{figure}[!ht]
   \centering
   \subfloat[\label{fig:conrot:skeleton-1a}]{
     \includegraphics[width=4.0cm]{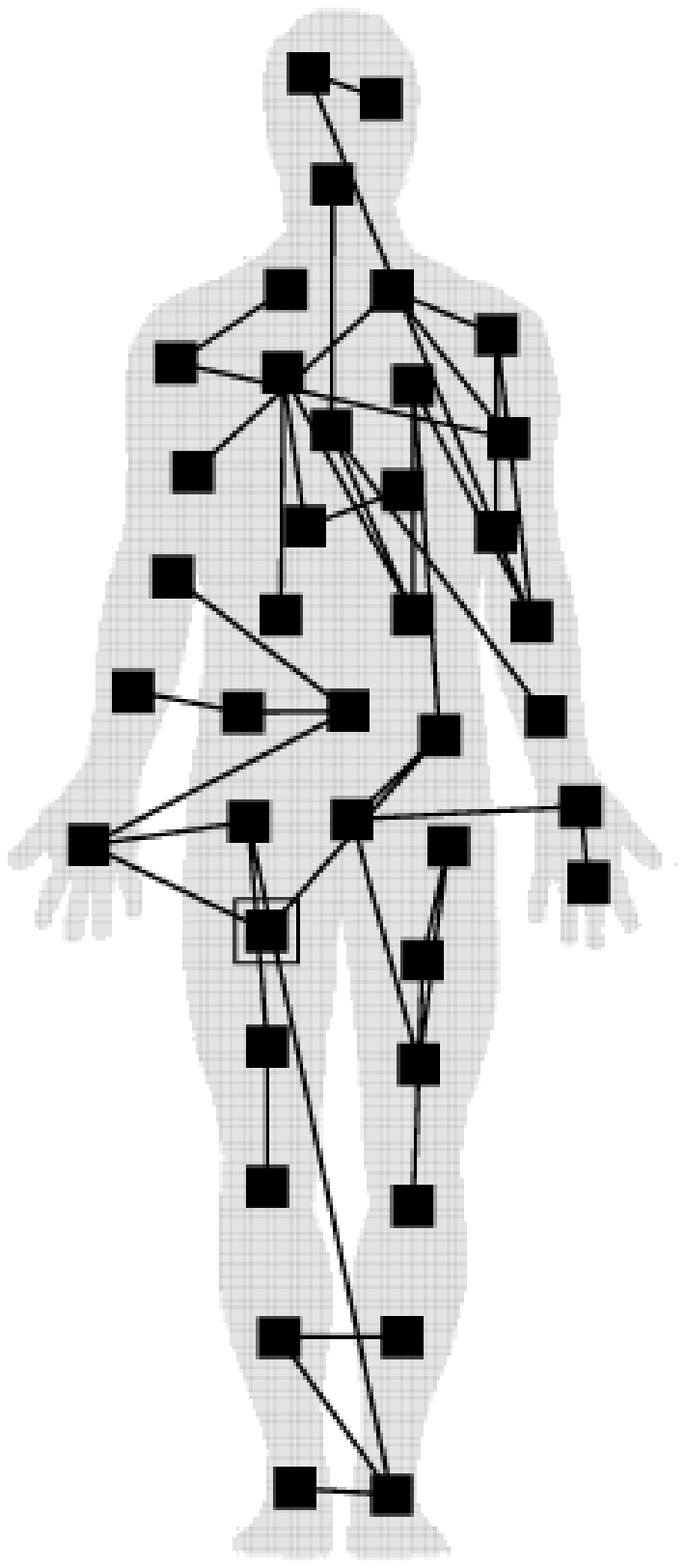}
   }
   \hspace{-15mm}
   \subfloat[\label{fig:conrot:skeleton-1b}]{
     \includegraphics[width=4.0cm]{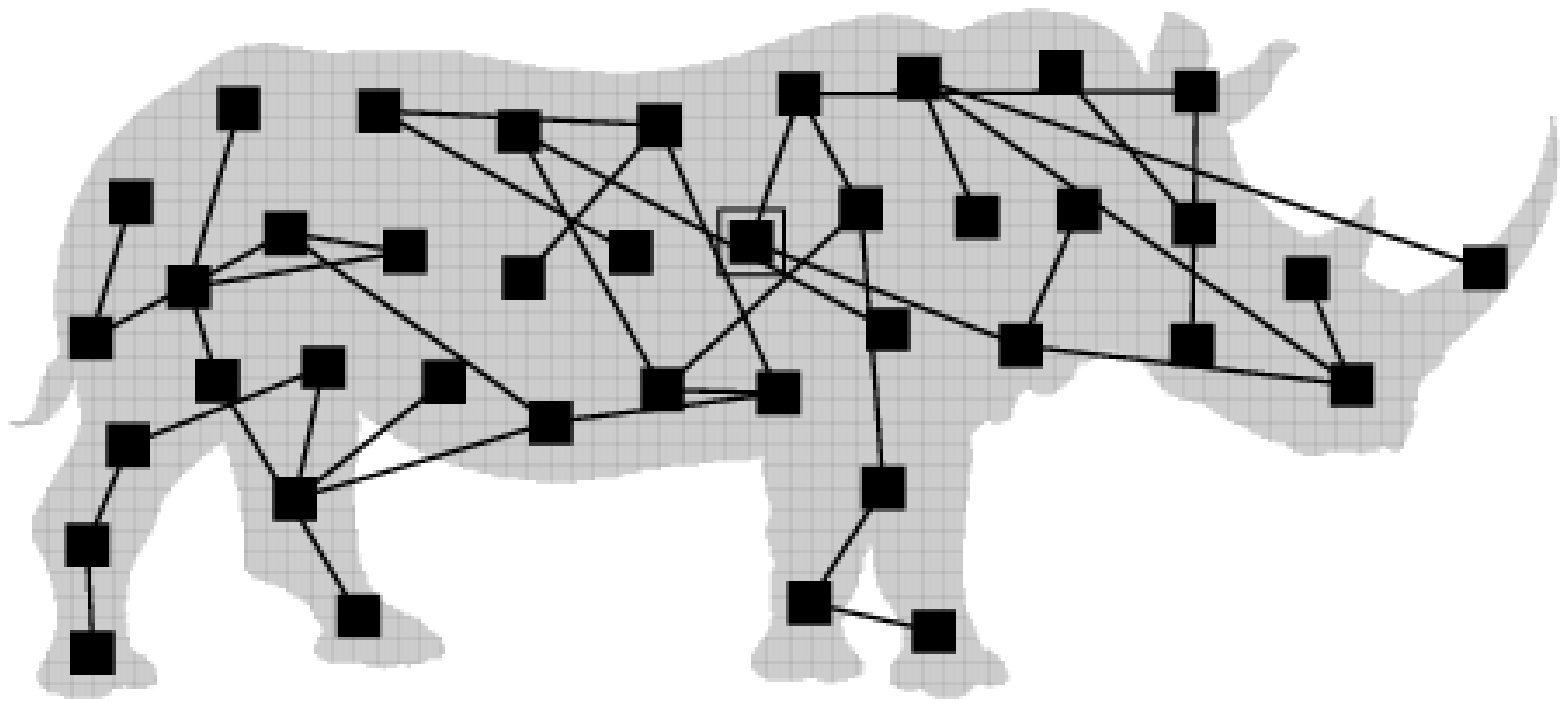}
   }
   \hspace{-15mm}
      \subfloat[\label{fig:conrot:skeleton-1c}]{
     \includegraphics[width=5cm]{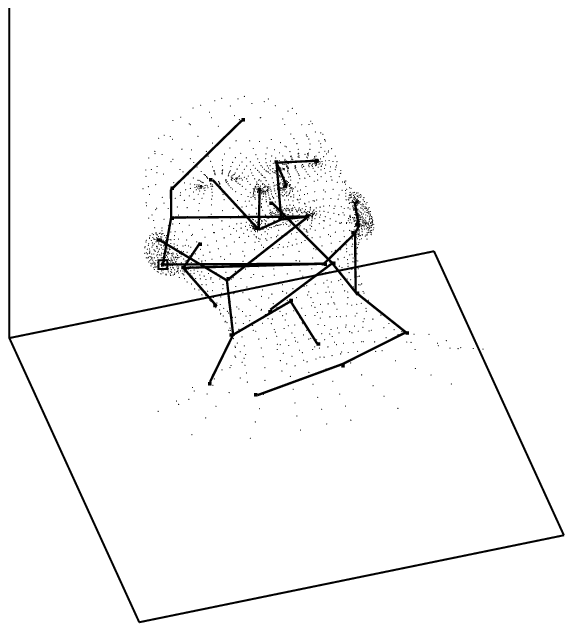}
   }
      \hspace{-15mm}
      \subfloat[\label{fig:conrot:skeleton-1d}]{
     \includegraphics[width=5cm]{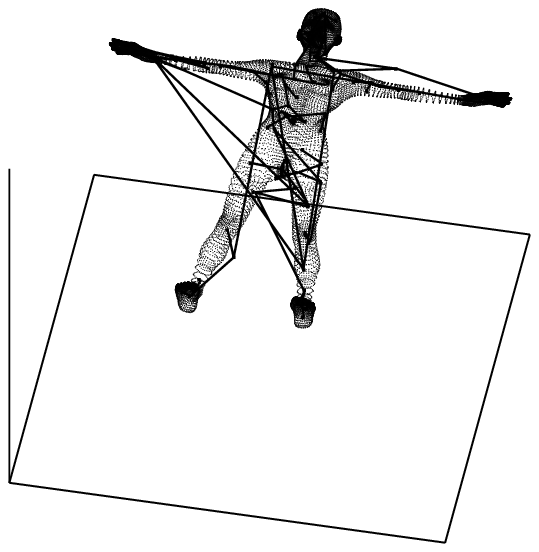}
   }
 \\
   \subfloat[\label{fig:conrot:skeleton-2a}]{
     \includegraphics[width=4.0cm]{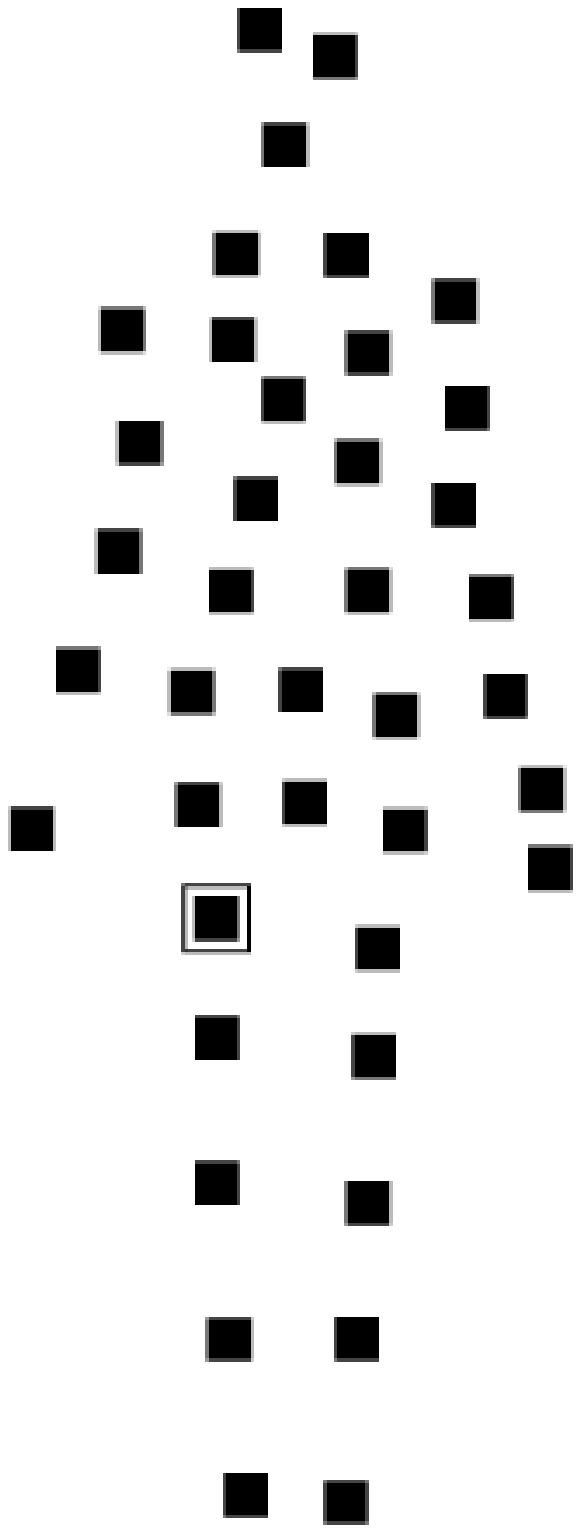}
   }
   \hspace{-15mm}
   \subfloat[\label{fig:conrot:skeleton-2b}]{
     \includegraphics[width=4.0cm]{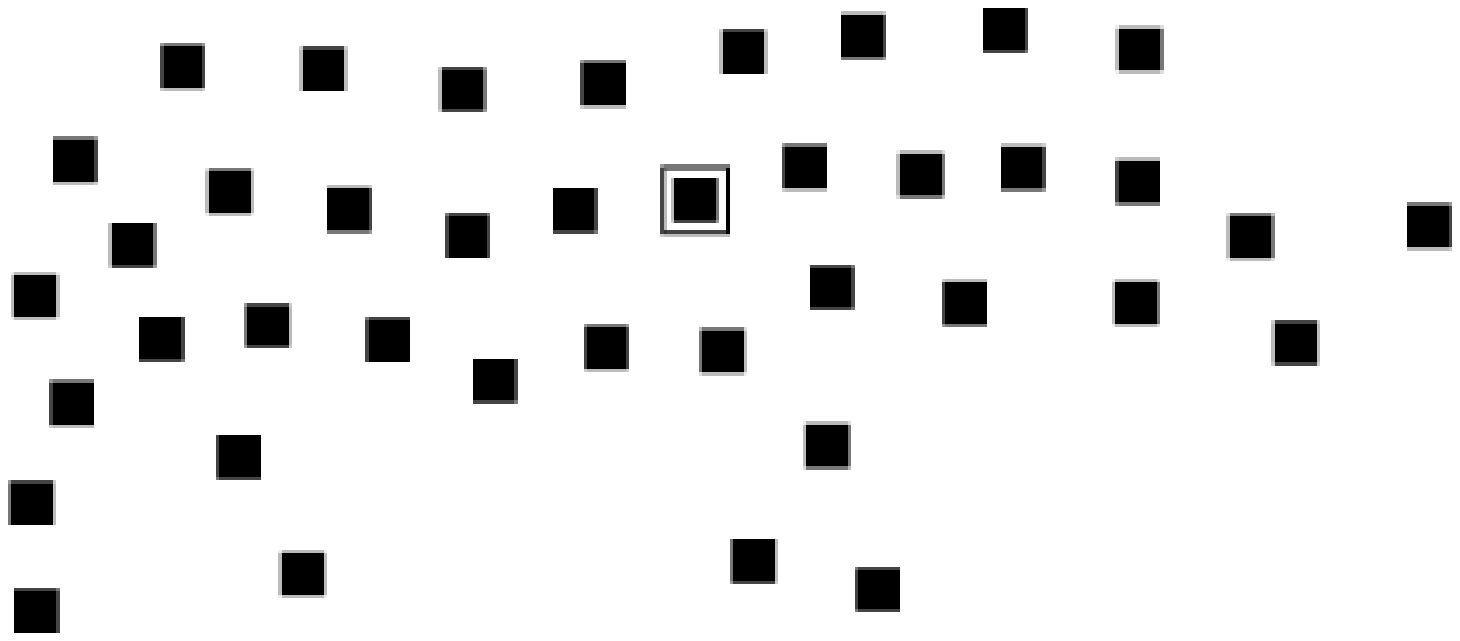}
   }
    \hspace{-15mm}
   \subfloat[\label{fig:conrot:skeleton-2c}]{
     \includegraphics[width=5cm]{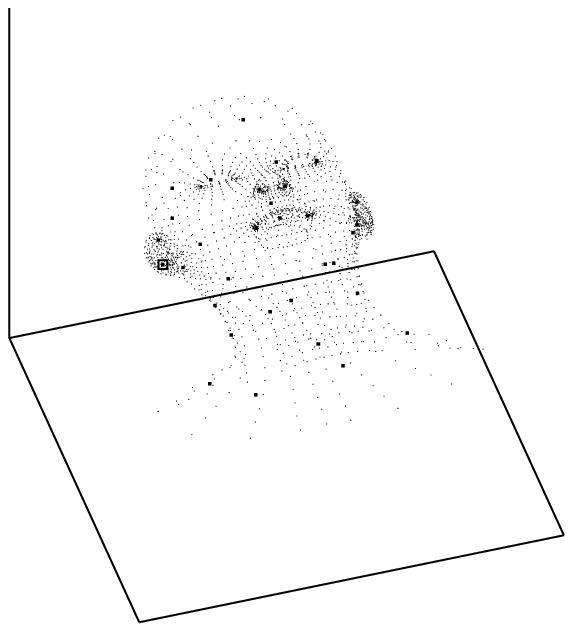}
   }
       \hspace{-15mm}
   \subfloat[\label{fig:conrot:skeleton-2d}]{
     \includegraphics[width=5cm]{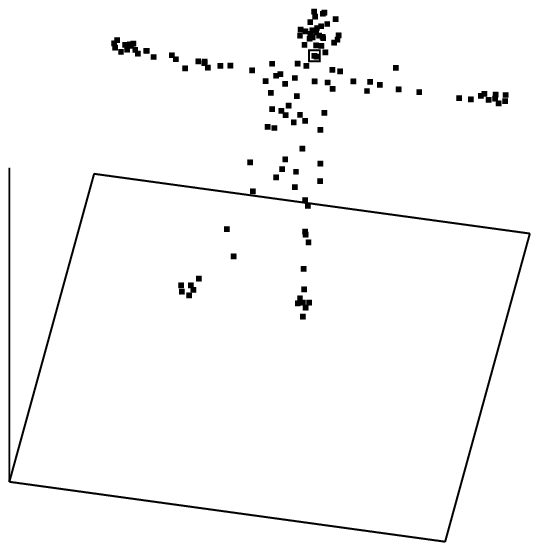}
   }
     \caption{\ac{TTOCONROT} effectively captures the fundamental structure of four objects in a 1-dimensional way. Figures \ref{fig:conrot:skeleton-1a}-\ref{fig:conrot:skeleton-1d} show the silhouette of a human, a rhinoceros, a 3d representation of a head, and a 3d representation of a woman, as well as the respective trees learned. Figures \ref{fig:conrot:skeleton-2a}-\ref{fig:conrot:skeleton-2d} show only the respective data points.}
   \label{fig:conrot:skeleton}
\end{figure}

Figure \ref{fig:conrot:skeleton} shows how the \ac{TTOCONROT} learned the skeleton of different objects in the 2-dimensional and the 3-dimensional domain. In all the cases the same schedule of parameters were used, and only the number of neurons employed was chosen proportionally to the number of data points contained in the respective data sets. It is important to remark that we did not invoke any post-processing of the edges, e.g., minimum spanning tree, and that the skeleton observed was exactly what our \ac{BSTSOM} learned. Firstly, Figures \ref{fig:conrot:skeleton-1a}-\ref{fig:conrot:skeleton-1d} illustrate the shapes of the silhouette of a human, a rhinoceros, a 3d representation of a head, and a 3d representation of a woman. The figures also show the trees learned from the respective data sets. Additionally figures \ref{fig:conrot:skeleton-2a}-\ref{fig:conrot:skeleton-2d} display only the data points, which in our opinion are capable of representing the fundamental structure of the four objects in a 1-dimensional way effectively.

As a final comment, we stress that all the shapes employed in the experiments involve the learning of the ``external'' structure of the objects. For the case of solid objects, if the internal data points are also provided, the \ac{TTOCONROT} is able to give an approximation of the so-called endo-skeleton, i.e., a representation in which the skeleton is built inside the solid object.

\subsection{Theoretical Analysis}
\label{sec:theoretical-analysis}

According to Kiviluoto \cite{Kiviluoto1996}, there are three different criteria for evaluating the quality of a map. The first criterion indicates how \textit{continuous} the mapping is, implying that input signals that are close (in the input space) should be mapped to codebooks that are close in the output space as well. A second criterion involves the \textit{resolution} of the mapping. Maps with high resolution possess the additional property that input signals that are distant in the input space should be represented by distant codebooks in the output space. A third criterion imposed on the accuracy of the mapping is aimed to reflect the \textit{probability distribution} of the input set. There exist a variety of measures for quantifying the quality of the topology preservation \cite{Arsuaga2005}. The author of \cite{Polzlbauer2004} surveys a number of relevant measures for the quality of maps, and these include the Quantization Error, the Topographic Product \cite{Bauer1992}, the Topographic Error \cite{Kiviluoto1996} and the Trustworthiness and Neighborhood Preservation \cite{Venna2001}. Although we are currently investigating \cite{AstudilloThesis} how the quality of any tree-based \ac{SOM} (not just our scheme) can be quantified using these metrics. The following arguments are pertinent.

The ordering of the weights (with respect to the position) of the neurons of the SOM has been proved for unidimensional topologies \cite{Conti1991,Kohonen1995,Rojas1996}. Extending these results to higher dimensional configurations or topologies leads to numerous unresolved problems. First of all, the question of what one means by ``ordering'' in higher dimensional spaces has to be defined. Further, the issue of the ``absorbing'' nature of the ``ordered state'' is open. Budinich, in \cite{Budinich1995}, explains intuitively the problems related to the ordering of neurons in higher dimensional configurations. Huang \textit{et al.}  \cite{Huang1998} introduce a definition of the ordering and show that even though the position of the codebook vectors of the SOM have been ordered, there is still the possibility that a sequence of stimuli will cause their disarrangement. Some statistical indexes of correlation between the measures of the weights and distances of the related positions have been introduced in \cite{Bauer1999}.


 With regard to the topographic product, the authors of \cite{Bauer1992} have shown the power of the metric by applying it on different artificial and real-world data sets, and also compared it with different measures to quantify the topology \cite{Bauer1999}. Their study concentrates on the traditional SOM, implying that the topologies evaluated were of a ``linear'' nature, with the consequential extension to 2-dimensions and 3-dimensions by means of grids only. In \cite{Haykin2008}, Haykin mention that the Topographic Product may be employed to compare the quality of different maps, even when these maps possess different dimensionality. However, he also noted that this measurement is only possible when the dimensionality of the topological structure is the same as the dimensionality of the feature space. Further, tree-like topologies were not considered in their study. To be more precise, most of the effort towards determining the concept of topology preservation for dimensions greater than unity are specifically focused on the SOM \cite{Bauer1992,Conti1991,Kohonen1995,Budinich1995,Huang1998,Bauer1999}, and do not define how a tree-like topology should be measured nor how to define the order in topologies which are not grid-based. Thus, we believe that even the tools to analyze the TTOCONROT are currently not available. The experimental results obtained in our paper, suggest that the TTOCONROT is able to train the NN so as to preserve the stimuli. However, in order to quantify the quality of this topology, the matter of defining a concept of ordering on tree-based structure has yet to be resolved. Although this issue is of great interest to us, this rather ambitious task lies beyond the scope of our present manuscript.

\section{Conclusions and Discussions}
\label{sec:conclusions}
\subsection{Concluding Remarks}
\acresetall 
In this paper, we have proposed a novel integration between the areas of \acp{ADS} and the \acp{SOM}. In particular we have shown how a tree-based \ac{SOM} can be adaptively transformed by the employment of an underlying \acf{BST} structure and subsequently, re-structured using rotations that are performed conditionally. These rotations on the nodes of the tree are local, can be done in constant time, and performed so as to decrease the \ac{WPL} of the entire tree. One of the main advantages of the algorithm, is that the user does not need to have \textit{a priori} knowledge about the topology of the input data set. Instead, our proposed method, namely the \ac{TTOCONROT}, infers the topological properties of the stochastic distribution, and at the same time, attempt to build the best \ac{BST} that represents the data set.

Incorporating the data structure's constraints in this ways has not being achieved by any of the related approaches included in the state-of-the-art. Our premise is that regions of the hyper-space that are accessed more often should be promoted to preferential spots in the tree representation, which yields to an improved stochastic representation.




As our experimental results suggest, the \ac{TTOCONROT} tree is indeed able to absorb the stochastic properties of the input manifold. It is also possible to obtain a tree configuration that can learn both, the stochastic properties in terms of access probabilities \textit{and} at the same time preserve the topological properties in terms of its skeletal structure.

\subsection{Discussions and Future Work}

As explained in Section \ref{sec:theoretical-analysis}, the work associated with measuring the topology preservation of the \ac{SOM}, including the proof of its convergence for the unidimensional case, has been performed for the traditional \ac{SOM} only.  
The questions are unanswered for
how a tree-like topology should be measured, and for defining the order in topologies which are not grid-based. Thus, we believe that even the tools for formally analyzing the \ac{TTOCONROT} are currently not available. The experimental results obtained in our paper, suggest that the \ac{TTOCONROT} is able to train the \ac{NN} so as to preserve the stimuli for which the concept of ordering on tree-based structures has yet to be resolved.

\CHECK
Even though our principal goal was to obtain a more accurate representation of the stochastic distribution, our results also suggest that the special configuration of the tree obtained by the \ac{TTOCONROT} can be further exploited so as to improve the time required for identifying the \ac{BMU}. The state-of-the-art includes different strategies that expand trees by inserting nodes (which can be a single neuron or a \ac{SOM}-layer) that essentially are based on a \ac{QE} measure. In some of these strategies, the error measure is based on the ``hits'', i.e., the number of times a neuron has been selected as the \ac{BMU}, which is, in principle, the same type of counter utilized by the \ac{CONROT}. Our strategy, \ac{TTOCONROT}, which asymptotically positions frequently accessed nodes close to the root, might incorporate a module, that taking advantage of the ``optimal'' tree and the \ac{BMU} counters already present in the \ac{TTOCONROT}, splits the node at the root level. Thus, the splitting operation will occur without the necessity of searching for the node with the largest \ac{QE}, under the assumption that a higher number of hits indicates that the degree of granularity of a particular neuron is lacking refinement.
The concept of using the root of the tree for growing a tree-based \ac{SOM} is itself pioneering, as far as we know, and the design and implementation details of this are currently being investigated.

\NORMAL



\bibliographystyle{plain}

\bibliography{TTO-CONROT_Jnl_fin,som-app}


\newcommand{\ACRO}{\acrodef}
\input{acronyms}

\end{document}

%% file: acronyms.tex
\ACRO{ABSTSOM}{Adaptive \acs{BSTSOM}} 
\ACRO{ADS}{Adaptive Data Structure}
\ACRO{AI}{Artificial Intelligence}
\ACRO{ART}{Adaptive Resonance Theory}
\ACRO{ANN}{Artificial Neural Network}
\ACRO{AUC}{Area Under the ROC Curve}
\ACRO{BMU}{Best Matching Unit}
\ACRO{BN}{Bayesian Network} 
\ACRO{BoA}{Bubble of Activity}
\ACRO{BST}{Binary Search Tree}
\ACRO{BSTSOM}{Binary Search Tree \acs{SOM}} 
\ACRO{CL}{Competitive Learning}
\ACRO{CM}{Confusion Matrix}
\ACRO{CONROT-BST}{Conditional Rotations for a \acs{BST}}
\ACRO{CONROT}{Conditional Rotations}
\ACRO{CPT}{Conditional Probabilities Table}
\ACRO{CS}{Computer Science}
\ACRO{CV}{Computer Vision}
\ACRO{DAG}{Directed Acyclic Graph}
\ACRO{DS}{Data Structure}
\ACRO{DT}{D-Tree}
\ACRO{ET}{Evolving Tree}
\ACRO{GCS}{Growing Cell Structures}
\ACRO{GG}{Growing Grid}
\ACRO{GHSOM}{Growing Hierarchical SOM}
\ACRO{GHTSOM}{Growing Hierarchical Tree SOM}
\ACRO{GNG}{Growing Neural Gas}
\ACRO{GT}{Growth Threshold}
\ACRO{GSOM}{Growing SOM}
\ACRO{HFM}{Hierarchical Feature Map}
\ACRO{HST}{Hyperplane Search Tree}
\ACRO{HSTSOM}{Hyperplane Search Tree \acs{SOM}}
\ACRO{IB}{Instance-Based}
\ACRO{IID}{Independently and Identically Distributed}
\ACRO{IGG}{Incremental Grid Growing}
\ACRO{IR}{Information Retrieval}
\ACRO{JPD}{Joint Probability Distribution}
\ACRO{KNN}[$k$-NN]{$k$-Nearest Neighbor}
\ACRO{KCONROT}[$k$-CONROT]{\acs{CONROT} for $k$-ary trees}
\ACRO{LVQ}{Learning Vector Quantization}
\ACRO{MAP}{\textit{Maximum A Posteriori}}
\ACRO{ML}{Machine Learning}
\ACRO{MLE}{Maximum Likelihood Estimation}
\ACRO{MQE}{Mean Quantization Error}
\ACRO{MST}{Minimum Spanning Tree}
\ACRO{MT}{Monotonic Tree}
\ACRO{NB}{Na\"{i}ve Bayes}
\ACRO{NG}{Neural Gas}
\ACRO{NN}{Neural Network}
\ACRO{PCA}{Principal Component Analysis}
\ACRO{PDF}{Probability Density Function}
\ACRO{PM}{Pattern Matching}
\ACRO{PR}{Pattern Recognition}
\ACRO{QE}{Quantization Error}
\ACRO{RHST}{Random Hyperplane Search Tree}
\ACRO{ROC}{Receiver Operating Characteristics}
\ACRO{SOM}{Self-Organizing Map}
\ACRO{SOTA}{Self-Organizing Tree Algorithm}
\ACRO{SOTM}{Self-Organizing Tree Map}
\ACRO{SVM}{Support Vector Machine}
\ACRO{TOD}{Threshold Order-Dependent}
\ACRO{TP}{Topographic Product}
\ACRO{TSVQ}{Tree-Structured VQ}
\ACRO{TSSOM}{Tree-Structured SOM}
\ACRO{TTOCONROT}{TTOSOM with Conditional Rotations}
\ACRO{TTOSOM}{Tree-based Topology Oriented SOM}
\ACRO{URL}{Uniform Resource Locator}
\ACRO{VQ}{Vector Quantization}
\ACRO{WDBC}{Wisconsin Diagnostic Breast Cancer}
\ACRO{WPL}{Weighted Path Length}

%% file: TTO-CONROT_Jnl_Fin.bbl
\begin{thebibliography}{10}

\bibitem{Adelson1962}
M.~Adelson-Velskii and M.~E. Landis.
\newblock An algorithm for the organization of information.
\newblock {\em Sov. Math. DokL}, 3:1259--1262, 1962.

\bibitem{Akram2013}
M.~U. Akram, S.~Khalid, and S.~A. Khan.
\newblock Identification and classification of microaneurysms for early
  detection of diabetic retinopathy.
\newblock {\em Pattern Recognition}, 46(1):107--116, 2013.

\bibitem{Alahakoon2000}
D.~Alahakoon, S.~K. Halgamuge, and B.~Srinivasan.
\newblock Dynamic self-organizing maps with controlled growth for knowledge
  discovery.
\newblock {\em IEEE Transactions on Neural Networks}, 11(3):601--614, 2000.

\bibitem{Allen1978}
B.~Allen and I.~Munro.
\newblock Self-organizing binary search trees.
\newblock {\em J. ACM}, 25(4):526--535, 1978.

\bibitem{Arsuaga2005}
E.~Arsuaga~Uriarte and F.~D\'{i}az~Mart\'{i}n.
\newblock Topology preservation in {SOM}.
\newblock {\em International Journal of Applied Mathematics and Computer
  Sciences}, 1(1):19--22, 2005.

\bibitem{AstudilloThesis}
C.~A. Astudillo.
\newblock {\em Self Organizing Maps Constrained by Data Structures}.
\newblock PhD thesis, Carleton University, 2011.

\bibitem{Astudillo2009a}
C.~A. Astudillo and B.~J. Oommen.
\newblock On using adaptive binary search trees to enhance self organizing
  maps.
\newblock In A.~Nicholson and X.~Li, editors, {\em 22nd Australasian Joint
  Conference on Artificial Intelligence (AI 2009)}, pages 199--209, 2009.

\bibitem{Astudillo2011TTOSOM}
C.~A. Astudillo and B.~J. Oommen.
\newblock Imposing tree-based topologies onto self organizing maps.
\newblock {\em Information Sciences}, 181(18):3798--3815, 2011.

\bibitem{Astudillo2013}
C.~A. Astudillo and B.~J. Oommen.
\newblock On achieving semi-supervised pattern recognition by utilizing
  tree-based {SOM}s.
\newblock {\em Pattern Recognition}, 46(1):293 -- 304, 2013.

\bibitem{Bauer1999}
H.~U. Bauer, M.~Herrmann, and T.~Villmann.
\newblock Neural maps and topographic vector quantization.
\newblock {\em Neural Networks}, 12(4-5):659 -- 676, 1999.

\bibitem{Bauer1992}
H.~U. Bauer and K.~R. Pawelzik.
\newblock Quantifying the neighborhood preservation of self-organizing feature
  maps.
\newblock {\em Neural Networks}, 3(4):570--579, July 1992.

\bibitem{Bitner1979}
J.~R. Bitner.
\newblock Heuristics that dynamically organize data structures.
\newblock {\em SIAM J. Comput.}, 8:82--110, 1979.

\bibitem{Blackmore1995}
J.~Blackmore.
\newblock Visualizing high-dimensional structure with the incremental grid
  growing neural network.
\newblock Master's thesis, University of Texas at Austin, 1995.

\bibitem{Budinich1995}
M.~Budinich.
\newblock On the ordering conditions for self-organizing maps.
\newblock {\em Neural Computation}, 7(2):284--289, 1995.

\bibitem{Carpenter1988}
G.~A. Carpenter and S.~Grossberg.
\newblock The art of adaptive pattern recognition by a self-organizing neural
  network.
\newblock {\em Computer}, 21(3):77--88, 1988.

\bibitem{Cheetham1993}
R.~P. Cheetham, B.~J. Oommen, and D.~T.~H. Ng.
\newblock Adaptive structuring of binary search trees using conditional
  rotations.
\newblock {\em IEEE Trans. on Knowl. and Data Eng.}, 5(4):695--704, 1993.

\bibitem{Conti1991}
P.~L. Conti and L.~De~Giovanni.
\newblock On the mathematical treatment of self organization: extension of some
  classical results.
\newblock In {\em Artificial Neural Networks - ICANN 1991, International
  Conference}, volume~2, pages 1089--1812, 1991.

\bibitem{Cormen2001}
T.~H. Cormen, C.~E. Leiserson, R.~L. Rivest, and C.~Stein.
\newblock {\em Introduction to Algorithms, Second Edition}.
\newblock McGraw-Hill Science/Engineering/Math, July 2001.

\bibitem{Datta1996}
A.~Datta, S.~M. Parui, and B.~B. Chaudhuri.
\newblock Skeletal shape extraction from dot patterns by self-organization.
\newblock {\em Pattern Recognition, 1996., Proceedings of the 13th
  International Conference on}, 4:80--84 vol.4, Aug 1996.

\bibitem{Deng2007}
D.~Deng.
\newblock Content-based image collection summarization and comparison using
  self-organizing maps.
\newblock {\em Pattern Recognition}, 40(2):718 -- 727, 2007.

\bibitem{Dittenbach2000}
M.~Dittenbach, D.~Merkl, and A.~Rauber.
\newblock The growing hierarchical self-organizing map.
\newblock In {\em Neural Networks, 2000. IJCNN 2000, Proceedings of the
  IEEE-INNS-ENNS International Joint Conference on}, volume~6, pages 15--19
  vol.6, 2000.

\bibitem{Dopazo1997}
J.~Dopazo and J.~M. Carazo.
\newblock Phylogenetic reconstruction using an unsupervised growing neural
  network that adopts the topology of a phylogenetic tree.
\newblock {\em Journal of Molecular Evolution}, 44(2):226--233, February 1997.

\bibitem{Duda2000}
R.~Duda, P.~E. Hart, and D.~G. Stork.
\newblock {\em Pattern Classification (2nd Edition)}.
\newblock Wiley-Interscience, 2000.

\bibitem{Fritzke1994}
B.~Fritzke.
\newblock {G}rowing {C}ell {S}tructures -- a self-organizing network for
  unsupervised and supervised learning.
\newblock {\em Neural Networks}, 7(9):1441--1460, 1994.

\bibitem{Fritzke1995a}
B.~Fritzke.
\newblock Growing {G}rid - a self-organizing network with constant neighborhood
  range and adaptation strength.
\newblock {\em Neural Processing Letters}, 2(5):9--13, 1995.

\bibitem{Fritzke1995}
B.~Fritzke.
\newblock A growing neural gas network learns topologies.
\newblock In G.~Tesauro, D.~S. Touretzky, and T.~K. Leen, editors, {\em
  Advances in Neural Information Processing Systems 7}, pages 625--632,
  Cambridge MA, 1995. MIT Press.

\bibitem{Guan2006}
L.~Guan.
\newblock Self-organizing trees and forests: A powerful tool in pattern
  clustering and recognition.
\newblock In {\em Image Analysis and Recognition, Third International
  Conference, ICIAR 2006, P{\'o}voa de Varzim, Portugal, September 18-20, 2006,
  Proceedings, Part I}, pages I: 1--14, 2006.

\bibitem{Haykin2008}
S.~Haykin.
\newblock {\em Neural Networks and Learning Machines}.
\newblock Prentice Hall, 3rd edition edition, 2008.

\bibitem{Huang1998}
G.~Huang, H.~A. Babri, and H.~Li.
\newblock Ordering of self-organizing maps in multi-dimensional cases.
\newblock {\em Neural Computation}, 10:19--24, 1998.

\bibitem{Kang2005}
H.-G. Kang and D.~Kim.
\newblock Real-time multiple people tracking using competitive condensation.
\newblock {\em Pattern Recognition}, 38(7):1045 -- 1058, 2005.

\bibitem{Kaplan2004}
H.~Kaplan.
\newblock {\em Handbook of Data Structures and Applications}, chapter 31:
  Persistent Data Structures, pages 31.1 -- 31.26.
\newblock Chapman and Hall/CRC, 2004.

\bibitem{Kaski1998}
S.~Kaski, J.~Kangas, and T.~Kohonen.
\newblock Bibliography of self-organizing map ({SOM}) papers: 1981--1997.
\newblock {\em Neural Computing Surveys}, 1:102--350, 1998.

\bibitem{Khosravi2008}
M.~H. Khosravi and R.~Safabakhsh.
\newblock Human eye sclera detection and tracking using a modified
  time-adaptive self-organizing map.
\newblock {\em Pattern Recognition}, 41(8):2571--2593, 2008.

\bibitem{Kiviluoto1996}
K.~Kiviluoto.
\newblock Topology preservation in self-organizing maps.
\newblock In P.~IEEE Neural~Networks Council, editor, {\em Proceedings of
  International Conference on Neural Networks, ICNN'96}, volume~1, pages
  294--299, New Jersey, USA, 1996.

\bibitem{Knuth1998}
D.~E. Knuth.
\newblock {\em The art of computer programming, volume 3: (2nd ed.) sorting and
  searching}.
\newblock Addison Wesley Longman Publishing Co., Inc., Redwood City, CA, USA,
  1998.

\bibitem{Kohonen1995}
T.~Kohonen.
\newblock {\em Self-Organizing Maps}.
\newblock Springer-Verlag New York, Inc., Secaucus, NJ, USA, 1995.

\bibitem{Koikkalainen1990}
P.~Koikkalainen and E.~Oja.
\newblock Self-organizing hierarchical feature maps.
\newblock {\em IJCNN International Joint Conference on Neural Networks},
  2:279--284, June 1990.

\bibitem{Lai1990}
T.~W.~H. Lai.
\newblock {\em Efficient maintenance of binary search trees}.
\newblock PhD thesis, University of Waterloo, Waterloo, Ont., Canada, 1990.

\bibitem{Liang2012}
Y.~Liang, M.~C. Fairhurst, and R.~M. Guest.
\newblock No titlea synthesised word approach to word retrieval in handwritten
  documents.
\newblock {\em Pattern Recognition}, 45(12):4225--4236, 2012.

\bibitem{Martinetz1991}
M.~Martinetz and K.~J. Schulten.
\newblock A ``neural-gas'' network learns topologies.
\newblock In {\em in Proceedings of International Conference on Articial Neural
  Networks}, volume~I, pages 397--402, North-Holland, Amsterdam, 1991.

\bibitem{Mehlhorn1979}
K.~Mehlhorn.
\newblock Dynamic binary search.
\newblock {\em SIAM Journal on Computing}, 8(2):175--198, 1979.

\bibitem{Merkl2003}
D.~Merkl, S.~Hui-He, M.~Dittenbach, and A.~Rauber.
\newblock Adaptive hierarchical incremental grid growing: An architecture for
  high-dimensional data visualization.
\newblock In {\em In Proceedings of the 4th Workshop on Self-Organizing Maps,
  Advances in Self-Organizing Maps}, pages 293--298, 2003.

\bibitem{Miikkulainen1990}
R.~Miikkulainen.
\newblock Script recognition with hierarchical feature maps.
\newblock {\em Connection Science}, 2(1\&2):83--101, 1990.

\bibitem{Ogniewicz1995}
R.~L. Ogniewicz and O.~K{\"u}bler.
\newblock Hierarchic voronoi skeletons.
\newblock {\em Pattern Recognition}, 28(3):343--359, 1995.

\bibitem{Oja2003}
M.~Oja, S.~Kaski, and T.~Kohonen.
\newblock Bibliography of self-organizing map ({SOM}) papers: 1998-2001
  addendum.
\newblock {\em Neural Computing Surveys}, 3:1--156, 2003.

\bibitem{Pakkanen2004}
J.~Pakkanen, J.~Iivarinen, and E.~Oja.
\newblock The {E}volving {T}ree --- a novel self-organizing network for data
  analysis.
\newblock {\em Neural Processing Letters}, 20(3):199--211, December 2004.

\bibitem{Peano1890}
G.~Peano.
\newblock Sur une courbe, qui remplit toute une aire plane.
\newblock {\em Mathematische Annalen}, 36(1):157--160, 1890.

\bibitem{Polla2009}
M.~P\"{o}ll\"{a}, T.~Honkela, and T.~Kohonen.
\newblock Bibliography of self-organizing map ({SOM}) papers: 2002-2005
  addendum.
\newblock Technical Report TKK-ICS-R23, Helsinki University of Technology,
  Department of Information and Computer Science, Espoo, Finland, December
  2009.

\bibitem{Polzlbauer2004}
G.~P\"olzlbauer.
\newblock Survey and comparison of quality measures for self-organizing maps.
\newblock In J\'an Parali\v{c}, Georg P\"olzlbauer, and Andreas Rauber,
  editors, {\em Proceedings of the Fifth Workshop on Data Analysis (WDA'04)},
  pages 67--82, Sliezsky dom, Vysok\'e Tatry, Slovakia, June 24--27 2004. Elfa
  Academic Press.

\bibitem{Rauber2002}
A.~Rauber, D.~Merkl, and M.~Dittenbach.
\newblock The {G}rowing {H}ierarchical {S}elf-{O}rganizing {M}ap: exploratory
  analysis of high-dimensional data.
\newblock {\em IEEE Transactions on Neural Networks}, 13(6):1331--1341, 2002.

\bibitem{Rojas1996}
R.~Rojas.
\newblock {\em Neural networks: a systematic introduction}.
\newblock Springer-Verlag New York, Inc., New York, NY, USA, 1996.

\bibitem{Samsonova2006}
E.~V. Samsonova, J.~N. Kok, and A.~P. {IJ}zerman.
\newblock Treesom: Cluster analysis in the self-organizing map.
\newblock {\em Neural Networks}, 19(6--7):935 -- 949, 2006.
\newblock Advances in Self Organising Maps - WSOM'05.

\bibitem{Singh2000}
R.~Singh, V.~Cherkassky, and N.~Papanikolopoulos.
\newblock Self-{O}rganizing {M}aps for the skeletonization of sparse shapes.
\newblock {\em Neural Networks, IEEE Transactions on}, 11(1):241--248, Jan
  2000.

\bibitem{Sleator1985}
D.~D. Sleator and R.~E. Tarjan.
\newblock Self-adjusting binary search trees.
\newblock {\em J. ACM}, 32(3):652--686, 1985.

\bibitem{Venna2001}
J.~Venna and S.~Kaski.
\newblock Neighborhood preservation in nonlinear projection methods: An
  experimental study.
\newblock In Georg Dorffner, Horst Bischof, and Kurt Hornik, editors, {\em
  ICANN}, volume 2130 of {\em Lecture Notes in Computer Science}, pages
  485--491. Springer, 2001.

\bibitem{Yao2000}
K.~C. Yao, M.~Mignotte, C.~Collet, P.~Galerne, and G.~Burel.
\newblock Unsupervised segmentation using a self-organizing map and a noise
  model estimation in sonar imagery.
\newblock {\em Pattern Recognition}, 33(9):1575 -- 1584, 2000.

\end{thebibliography}
